\documentclass[11pt]{article}

\PassOptionsToPackage{sort}{natbib}

\usepackage[final]{acl}

\usepackage{times}
\usepackage{latexsym}
\usepackage{graphicx}
\usepackage{booktabs}
\usepackage{multirow}
\usepackage{amsmath}
\usepackage{siunitx}
\usepackage{subcaption}
\usepackage{amsfonts}
\usepackage{makecell}
\usepackage[most]{tcolorbox}

\usepackage[T1]{fontenc}
\usepackage[utf8]{inputenc}
\usepackage{microtype}
\usepackage{inconsolata}

\usepackage[dvipsnames]{xcolor}
\definecolor{ForestGreen}{rgb}{0.13, 0.55, 0.13}

\title{Cross-lingual Calibration of Pre-Generation Success Probes for Multilingual LLM Routing}

\author{
 \textbf{Andrea Paganelli\textsuperscript{1,2}},
 \textbf{Stefano Civelli\textsuperscript{1}},
 \textbf{Pietro Bernardelle\textsuperscript{1}},
 \textbf{Gianluca Demartini\textsuperscript{1}}
\\
 \textsuperscript{1}The University of Queensland
 \textsuperscript{2}Polytechnic University of Milan
\\
 \small{
   \textbf{Correspondence:} \href{mailto:s.civelli@uq.edu.au}{s.civelli@uq.edu.au}
 }
}

\begin{document}
\maketitle

\begin{abstract}
Pre-generation success probes estimate response correctness from a language model's hidden activations before decoding, enabling cost-aware routing. While prior work has demonstrated their utility primarily on English inputs, we study their reliability across languages along three dimensions: \textbf{(1)} whether they preserve the ranking of likely successes and failures (\textsc{discrimination}); \textbf{(2)} whether they retain probabilities that match observed success frequencies (\textsc{calibration}); and \textbf{(3)} whether they produce scores comparable enough across candidate models for cost-aware multilingual routing (\textsc{utility}). Using 3,000 MATH problems in 10 languages and 8 open-weight model configurations, we compare cross-lingual transfer from English-trained probes and equal-budget pooled multilingual probes. English-trained probes retain useful cross-lingual discrimination but become less well calibrated after transfer. Pooled multilingual supervision improves both properties and yields more reliable estimates of success. In routing experiments, the pooled router achieves a 0.7\% higher test success rate while reducing modeled cost by 13.0\% relative to always selecting the model with the highest average success. These results show that multilingual routing requires success estimates that remain well calibrated and comparable across languages and models.
\end{abstract}

\section{Introduction}
\label{sec:intro}
\begin{figure}[t]
  \centering
  \includegraphics[width=\linewidth]
  {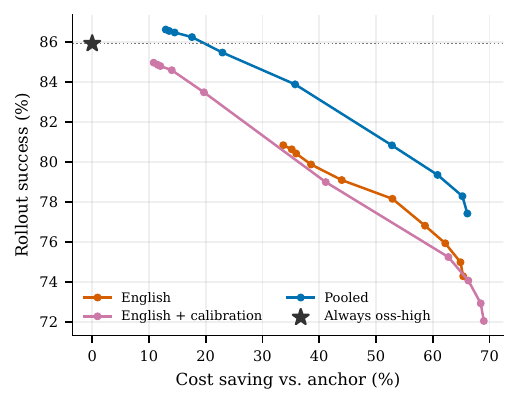}
  \caption{
    Multilingual routing with pre-generation success probes. Test rollout success versus modeled cost saving relative to always selecting \texttt{gpt-oss-high}, on the balanced ten-language test set. Each point corresponds to a different value of the routing trade-off coefficient $\lambda$.
    Curves compare English-trained, recalibrated English-trained, and equal-budget pooled multilingual probes used for routing.
    }
  \label{fig:routing-frontiers}
\end{figure}

Large language models (LLMs) do not perform consistently across languages. A model may solve a mathematical problem in English and fail on a semantically equivalent formulation in another language \citep{shi2022languagemodelsmultilingualchainofthought,ahuja-etal-2023-mega,ko-etal-2025-understand,son-etal-2025-linguistic}. At the same time, always using the strongest available model is costly and often unnecessary, since a cheaper model may be equally capable of solving the same problem \citep{chen2023frugalgpt,aggarwal2024automix,chen2024routerdc,ong2024routellm}. Cost-aware model routing aims to address this trade-off by selecting, for each input, the least expensive model likely to produce a correct response. Such a decision requires estimating, before decoding, how likely each candidate model is to answer correctly.

Recent work has shown that a model's hidden activations contain information about the correctness of its future response \citep{azaria2023internal,lugoloobi2026failures}. Linear probes can extract this information after the prompt has been processed but before the first output token is generated. These \emph{pre-generation success probes} can therefore support routing without first generating and evaluating a candidate response \citep{di2026code,lugoloobi2026failures,varshney2026router}. However, existing evidence is largely limited to English inputs, leaving unclear whether these probes remain reliable across languages and whether their estimates are sufficiently comparable for multilingual routing. We therefore ask: \textbf{(RQ1)} How reliably do pre-generation success probes transfer from English to non-English target languages? \textbf{(RQ2)} How effectively do the resulting success estimates support cost-aware routing?

We investigate these questions using 3,000 problems from the MATH dataset \citep{hendrycks2021math}, translated from English into 9 additional languages, and 8 open-weight model configurations. For each problem, language, and model configuration, we collect $n = 5$ sampled generations to estimate empirical success and extract residual-stream activations at every model depth before decoding. We train layer-wise ridge-regression probes under \textit{monolingual} training and equal-budget \textit{pooled multilingual} training. We then compare English-to-target cross-lingual transfer with pooled multilingual supervision, using a probe trained directly on labelled data in the target language as an oracle reference.
We assess probes' reliability through three increasingly demanding requirements: \textbf{(1)} probes should preserve the ordering of likely successes and failures (\textsc{discrimination}); \textbf{(2)} their predicted probabilities should remain aligned with observed success frequencies (\textsc{calibration}); and \textbf{(3)} their scores should be sufficiently comparable across candidate models to support effective cost-aware choices (\textsc{utility}).

Our results show that English-trained probes retain substantial cross-lingual discrimination, particularly at intermediate layers \citep{civelli2026geometry}. Their scores, however, do not retain a consistent probabilistic meaning across languages.  Pooled multilingual supervision reduces expected calibration error by 38--69\% across models and produces a stronger success--cost frontier in routing (see Figure \ref{fig:routing-frontiers}). Overall, our findings show that preserving cross-lingual rankings is only one requirement for reliable multilingual routing. Effective routing also requires success probabilities whose scale remains meaningful across languages and sufficiently comparable across candidate models.

\section{Related Work}
\label{sec:related}

\paragraph{Pre-generation prediction from  activations.}
Prior work has investigated whether a model's internal state contains information about properties of its eventual response before generation begins. Early studies identify activation patterns associated with truthfulness and correctness \citep{azaria2023internal}. Related work shows that human-annotated problem difficulty is linearly decodable from pre-generation representations \citep{lugoloobi2025difficulty, civelli2026geometry}. Moving from these general response and problem-level properties to model-specific success, \citet{cencerrado2025no} estimate forthcoming answer correctness from question-only activations by constructing correctness directions from the difference between the mean activations of successful and unsuccessful examples. These directions transfer well across factual knowledge tasks, but their predictive performance is weaker on mathematical reasoning. \citet{lugoloobi2026failures} extend this line of work by training supervised linear probes directly on labelled successes and failures. They show that these probes can predict model success on mathematical and coding tasks and, crucially, use the resulting estimates for cost-aware model routing. Subsequent approaches similarly exploit prefilling-time signals for self-evaluation \citep{kasnavieh2026introlm} or use an open-weight model's activations to predict the correctness of separate target models for routing \citep{varshney2026router}. Together, these studies establish that pre-generation representations encode signals correlated with correctness and that these signals can support routing decisions. 

\paragraph{Cross-lingual transfer of internal signals.}
Multilingual models have been shown to map semantically equivalent inputs in different languages to partially aligned representation spaces \cite{bernardelle2026llms,kim2025language,li2025exploring,schut2025multilingual,wendler2024llamas}, enabling zero-shot cross-lingual transfer \citep{pires2019multilingual,conneau2020unsupervised,libovicky2020neutrality}. Nevertheless, multilingual representation differences in tokenisation, pretraining data, and language-specific model competence can produce substantial variation across languages \citep{rust2021tokenizer}. Moreover, representational alignment alone does not guarantee that task-relevant information will transfer. In a controlled multilingual setting, \citet{hua2024mothello} show that strongly aligned representations can coexist with poor cross-lingual transfer, demonstrating that shared geometry is not sufficient for transferring a learned predictor. This distinction is particularly important for meta-cognitive signals. Concurrent work finds that problem difficulty has a partly shared cross-lingual geometry, especially in intermediate model layers \citep{civelli2026geometry}. Similarly, \citet{datta2026hallucination} report that activation-based hallucination detectors trained in English deteriorate under zero-shot cross-lingual transfer and benefit from multilingual supervision. 

\paragraph{Calibration under distributional shifts.}
Discrimination and calibration describe complementary properties of a predictor. A predictor may rank successful examples above unsuccessful ones while assigning scores that do not correspond to empirical outcome frequencies \citep{guo2017calibration,zadrozny2002transforming,naeini2015obtaining,platt1999probabilistic}. 
Prior work has shown that cross-lingual transfer can substantially degrade the calibration of output-level predictions \citep{jiang2022calibrating}. More recently, \citet{zhou2025beyond} study multilingual confidence signals in internal representations and find that intermediate layers can provide better-calibrated confidence estimates than final-layer representations, particularly outside English. This result connects multilingual calibration to representational depth, but does not establish whether supervised pre-generation success probes preserve calibration when transferred between languages or whether multilingual supervision improves their downstream routing utility.

\paragraph{Cost-aware LLM routing.}
LLM routing and cascading seek to reduce inference cost by assigning each query to the least expensive model expected to answer it correctly \citep{chen2023frugalgpt,ong2024routellm,ding2024hybrid,dekoninck2025unified,jitkrittum2025universal}. Most routing methods estimate model suitability from external signals such as query embeddings, preference data, previously observed outputs, or uncertainty measures. Activation-based routers instead use internal representations computed during prompt prefilling to estimate model-specific success before generation begins \citep{lugoloobi2026failures,varshney2026router}. For routing, however, predictive ranking alone may be insufficient. Model scores must be comparable across candidate models and operating conditions if they are used to set thresholds or optimise expected success under a cost constraint. 

Existing activation-based routing work has not examined whether success estimates or routing thresholds remain valid under language shift.

\section{Methodology}
\label{sec:methodology}

\subsection{Formulation}
\label{sec:task}
Let $x_i$ denote a mathematical problem, $\ell$ the language in which it is presented, and $m$ an evaluated language model. We write $x_i^{(\ell)}$ for the corresponding language-specific prompt. Let $\pi_m(r \mid x_i^{(\ell)})$ denote the probability that model $m$, under the fixed decoding policy, produces response $r$ when given $x_i^{(\ell)}$. Accordingly, $\pi_m(\cdot \mid x_i^{(\ell)})$ denotes the full distribution over possible responses.
We write $R_{m,\ell,i} \sim \pi_m(\cdot \mid x_i^{(\ell)})$ to denote one response sampled from this distribution. The corresponding single-generation success probability is
\begin{equation}
    p_{m,\ell,i}
    =
    \Pr\left[
    \operatorname{Correct}(R_{m,\ell,i})=1
    \right].
    \label{eq:success-probability}
\end{equation}

Following \citet{lugoloobi2026failures}, for every model--problem--language tuple, we sample $n=5$ independent responses and define
\begin{equation*}
    k_{m,\ell,i}
    =
    \sum_{r=1}^{n}
    \mathbb{I}\left[
    \operatorname{Correct}
    \left(R^{(r)}_{m,\ell,i}\right)=1
    \right]
\end{equation*}
as the number of correct responses. The empirical success score is therefore defined as
\begin{equation}
    y_{m,\ell,i}
    =
    \frac{k_{m,\ell,i}}{n}
    \in
    \{0,0.2,0.4,0.6,0.8,1\}.
    \label{eq:success}
\end{equation}
This score estimates the probability in Equation~\ref{eq:success-probability}, since $\mathbb{E}[y_{m,\ell,i}]=p_{m,\ell,i}$.

Since $p_{m,\ell,i}$ is conditional on both the model and the language of the prompt, the same underlying problem can have different success probabilities across models and translations. 


\subsection{Data}
\label{sec:data}

We sample 3,000 English-language problems from the MATH dataset \citep{hendrycks2021math} and translate each into Arabic, Chinese, French, Italian, Russian, Swahili, Telugu, Thai, and Turkish using DeepL \citep{deepl2026}.
To preserve the mathematical content and structure of the source problems, we protect mathematical notation and structured blocks during translation. We apply regular-expression checks to detect malformed notation, and use COMET-Kiwi \citep{rei2022cometkiwi} as a reference-free estimate of translation adequacy. For further details on translation quality see Appendix~\ref{a:translation}.

We partition the 3,000 underlying MATH problems into training, validation, and test sets using a 70/15/15 split. We split by underlying problem rather than by translated prompt, so all ten language-specific versions of a problem remain in the same partition, with the resulting assignment shared across all models. The training split is used to fit probe parameters, the validation split is used for regularisation, reporting-layer, recalibration, and routing-policy selection, and the test split is reserved for final evaluation.

\subsection{Models and Generation}
\label{sec:models}

Our experiments include 8 configurations of instruction-tuned \cite{ouyang2022training}, open-weight models: \texttt{Qwen3} at \texttt{0.6B}, \texttt{1.7B}, \texttt{4B}, and \texttt{8B} parameters \citep{yang2025qwen3}; \texttt{Qwen3.5} at \texttt{4B} and \texttt{9B} parameters \citep{qwen3.5}; and \texttt{gpt-oss-20b} \citep{openai2025gptoss120bgptoss20bmodel}, evaluated under low- and high-reasoning settings, denoted \texttt{gpt-oss-low} and \texttt{gpt-oss-high}, respectively.

To simulate a realistic usage setting, each model receives only the mathematical problem, formatted using its native chat template, with no additional instructions. For each model--problem--language tuple, we sample $n=5$ responses at a temperature of 0.7, extract the final  answer, and aggregate the resulting binary correctness labels according to Equation~\ref{eq:success}.

\subsection{Layer-wise Representation Extraction}
\label{sec:representations}

For each model--problem--language tuple, we perform a single forward pass using the Hugging Face Transformers library \citep{wolf-etal-2020-transformers} and extract residual-stream activations at every model depth. For a model with $L$ transformer blocks, we consider $L+1$ representation depths. Depth $0$ denotes the initial residual-stream state obtained by mapping the input tokens through the model's embedding matrix, before any transformer block is applied. Depths $j=1,\ldots,L$ denote the residual-stream states after the corresponding transformer blocks.

At each depth, we record the activation $\mathbf{h}^{(j)}_{m,\ell,i} \in \mathbb{R}^{d_m}$ at the final prompt token, before the generation of the first output token, where $d_m$ is the hidden dimension of model $m$.\footnote{We select the final prompt token because it can attend to the entire input, making its activation a representation of the model state from which generation begins.} 

\subsection{Success-Probes Training and Selection}
\label{sec:probes}

\paragraph{Probes training.}

Using the layer-wise representations defined above, we train a separate linear ridge-regression probe for each model, layer, and supervision regime.
Under \textbf{(1)} \emph{monolingual supervision}, the probe is trained on the complete training set in a single language, whereas under \textbf{(2)} \emph{pooled multilingual supervision}, the same overall training budget is distributed evenly across the ten languages. 

Before probe training, each activation dimension is standardised by removing its mean and scaling it to unit variance. Both statistics are estimated exclusively from the corresponding training split.
The probe produces the raw prediction $\widehat{p}^{(j),\mathrm{raw}}_{m,\ell,i} = \mathbf{w}^{\top} \widetilde{\mathbf{h}}^{(j)}_{m,\ell,i} + b,$ where $\widetilde{\mathbf{h}}^{(j)}_{m,\ell,i}$ is the standardised activation. 
For each candidate ridge coefficient $\alpha\in\{10,10^2,10^3,10^4,10^5\}$ we fit a separate set of probe parameters $\mathbf{w}$ and $b$ on the training split by minimizing
\begin{equation*}
    \sum_i
    \left(
    y_{m,\ell,i}
    -\widehat{p}^{(j),\mathrm{raw}}_{m,\ell,i}
    \right)^2
    +
    \alpha\lVert\mathbf{w}\rVert_2^2.
    \label{eq:ridge}
\end{equation*}

Thus, $\mathbf{w}$ and $b$ are optimized for a fixed value of $\alpha$; the value of $\alpha$ itself is subsequently selected using the validation procedure described in the next paragraph.
To constrain the regression output, we project its predictions onto $[0,1]$:
$
    \widehat{p}^{(j)}_{m,\ell,i}
    =
    \min
    \left\{
    1,
    \max
    \left\{
    0,
    \widehat{p}^{(j),\mathrm{raw}}_{m,\ell,i}
    \right\}
    \right\}.
    \label{eq:probe-projection}
$
This projection is applied when computing validation and test metrics and when using predictions for recalibration or routing. 



\paragraph{Validation-based probe selection.}

We use validation Brier score to select the ridge coefficient and reporting layer because the probes are intended to estimate single-generation success probabilities. For an evaluation set $\mathcal{D}$, let $\widehat{p}_i$ denote the projected predicted probability, $k_i$ the number of successful rollouts, and $n_i$ the total number of retained rollouts for problem $i$. We compute the rollout-level Brier score as
\begin{equation}
\operatorname{BS}(\mathcal{D})
=
\frac{
\sum_{i\in\mathcal{D}}
\left[
k_i(1-\widehat{p}_i)^2
+
(n_i-k_i)\widehat{p}_i^2
\right]
}{
\sum_{i\in\mathcal{D}} n_i
}.
\label{eq:brier}
\end{equation}
In our experiments, $n_i=5$ for every problem. This score is the mean squared error between the predicted probability and the individual binary rollout outcomes, and therefore evaluates the same single-generation success probability defined in Equation~\ref{eq:success-probability}. It summarizes overall probabilistic quality and may improve through better discrimination, better calibration, or both.

Selection proceeds in two stages. First, as stated in the previous paragraph, we fit one probe for each candidate
$
\alpha\in\{10,10^2,10^3,10^4,10^5\}
$
on the training split for every model, layer, and supervision regime, and select the coefficient that minimizes validation Brier score:
\begin{equation*}
\alpha^\star_{m,j,s}
=
\arg\min_{\alpha}
\operatorname{BS}^{\mathrm{val}}_{m,j,s}(\alpha),
\label{eq:alpha-selection}
\end{equation*}
where $s$ denotes the supervision regime. For a monolingual probe, including the English probe and each same-language reference probe, the candidate coefficients are evaluated on validation examples in the probe's training language. For the pooled multilingual probe, they are evaluated on the union of the validation sets from all ten languages. 
The resulting layer-wise ridge-coefficient selections are reported in Appendix~\ref{app:alpha-selection}.

Second, after selecting $\alpha^\star_{m,j,s}$ at every layer, we select one shared reporting layer for each model. Using a different layer for each supervision regime or evaluation metric could selectively favour individual conditions. For
$
s\in{\mathrm{English},\mathrm{Pooled}},
$
we therefore define
\begin{equation*}
    \operatorname{macroBS}^{\mathrm{val}}_{s,m,j}
    =
    \frac{1}{|\mathcal{L}_{\mathrm{target}}|}
    \sum_{\ell\in\mathcal{L}_{\mathrm{target}}}
    \operatorname{BS}^{\mathrm{val}}_{s,m,j,\ell},
\label{eq:macro-brier}
\end{equation*}
where each score is computed using the corresponding validation-selected coefficient $\alpha^\star_{m,j,s}$, and $\mathcal{L}*{\mathrm{target}}$ contains the nine non-English target languages. We select
\begin{equation}
\begin{aligned}
    j_m^\star
    &=
    \arg\min_j
    \left[
    \frac{1}{2}
    \operatorname{macroBS}^{\mathrm{val}}_{\mathrm{English},m,j}
    \right. \\
    &\qquad\left.
    {}+
    \frac{1}{2}
    \operatorname{macroBS}^{\mathrm{val}}_{\mathrm{Pooled},m,j}
    \right].
\end{aligned}
\label{eq:shared-layer}
\end{equation}
Same-language probes do not influence the layer choice. Once selected, $j_m^\star$ is frozen across all evaluation regimes (Section~\ref{sec:evaluation}) and used for all scalar test metrics and downstream analyses.

\subsection{Cross-lingual Probe Reliability}
\label{sec:evaluation}
To answer \textbf{RQ1}, we evaluate probe quality on the same held-out examples from each of the nine non-English target languages. Excluding English ensures that all reported comparisons measure performance under language transfer.

We compare three regimes: \textbf{(a) English transfer (\textsc{Eng.})}, where a probe trained under \textit{monolingual supervision} on English is applied to the target language; \textbf{(b) pooled multilingual supervision (\textsc{Pool})}, where a probe trained under \textit{pooled multilingual supervision} is evaluated separately on each target language; and \textbf{(c) same-language reference (\textsc{Oracle})}, where a probe trained under \textit{monolingual supervision} in the target language is evaluated on that language. The oracle therefore uses labelled target-language training data and serves as a reference for the other two conditions.

\paragraph{Discrimination.}
We evaluate discrimination using rollout-weighted AUROC. A problem with $k_i$ successful and $n_i-k_i$ failed generations contributes its score $\widehat{p}_i$ with weight $k_i$ to the positive class and weight $n_i-k_i$ to the negative class. AUROC therefore measures the probability that a randomly selected successful rollout receives a higher score than a randomly selected failed rollout.

AUROC depends only on the ordering of predictions and is invariant to strictly monotonic transformations of their values. A probe can therefore achieve high AUROC while being systematically overconfident or underconfident.

\paragraph{Calibration.}
Calibration measures whether predicted probabilities agree with observed success frequencies. 
We diagnose this using sample-weighted Expected Calibration Error (ECE) with ten equal-width bins and reliability diagrams. ECE summarizes the average absolute difference between predicted and observed success rates across bins, while reliability diagrams show where these differences occur. 

Calibration does not measure whether a probe meaningfully separates easy and difficult problems. For example, a probe that always predicts the overall success rate may be perfectly calibrated while having no discriminative ability. 


\paragraph{Probabilistic quality.}
We evaluate the overall quality of the predicted success probabilities using the rollout-level Brier score in Equation~\ref{eq:brier} and aggregated binomial negative log-likelihood (NLL). Both are strictly proper scoring rules: in expectation, they are minimized by reporting the true success probability. They jointly reward predictions that are correctly scaled and informative.
Brier score applies a quadratic penalty to probability errors and is bounded, whereas NLL applies a logarithmic penalty and therefore penalizes confidently incorrect predictions more strongly. 

\paragraph{}
Neither metric isolates calibration from discrimination: an improvement may result from a better probability scale, a better ordering of examples, or both. We consequently use AUROC to measure discrimination, calibration diagnostics to evaluate probability scale, and Brier score and NLL to summarize their combined effect on probability quality (more details can be found in Appendix~\ref{a:metric-differences}).

\subsection{Multilingual Routing Utility}
\label{sec:routing-method}
To answer \textbf{RQ2}, we evaluate routing over a language-balanced mixture of all ten languages. This scope differs deliberately from the nine-language probe-quality evaluation since in a real routing setting prompts may come from any of the supported languages, including English.
We use the predicted success probabilities to select a target model before generation. Given a set of candidate models $\mathcal{M}$, the router chooses
\begin{equation}
m^\star(x)
=
\arg\max_{m\in\mathcal{M}}
\left[
\widehat{p}_{m}(x)
-
\lambda c_m
\right],
\label{eq:routing}
\end{equation}
where $c_m$ is the estimated inference cost (refer to Appendix~\ref{a:routing-cost}) and $\lambda$ controls the success--cost trade-off. 
We evaluate the fixed grid
$
\lambda \in
\{0,\ .01,\ .02,\ .05,\ .10,\ .20,\ .35,\ .50,\ .75,\ 1\},
$
with each value defining one operating point on the success--cost frontier.


Our main analysis uses encoder--target decoupled routing (ETD)~\cite{varshney2026router}, in which a fixed encoder supplies the representation used to predict the success of every candidate target model. We use \texttt{Qwen3-4B} as the shared encoder (refer to Appendix \ref{a:encoder-selection}) and train a separate target-specific ridge head for each candidate model at every encoder layer. 
We select one ETD reporting layer using the joint validation-Brier rule in Equation~\ref{eq:shared-layer}, additionally macro-averaged over the 8 target models. Predictions from this frozen layer are used in all subsequent recalibration and routing conditions. 

\newcommand{\ETDSelectedLayer}{23}
\newcommand{\MeanSelfEnglishAUROC}{0.697}
\newcommand{\MeanSelfPooledAUROC}{0.724}
\newcommand{\MeanSelfOracleAUROC}{0.746}
\newcommand{\MeanSelfEnglishBrier}{0.171}
\newcommand{\MeanSelfPooledBrier}{0.154}
\newcommand{\MeanETDEnglishAUROC}{0.684}
\newcommand{\MeanETDPooledAUROC}{0.734}
\newcommand{\MeanETDEnglishBrier}{0.181}
\newcommand{\MeanETDCalibratedBrier}{0.169}
\newcommand{\MeanETDPooledBrier}{0.152}
\newcommand{\CalibrationGapClosedPercent}{26\%}
\newcommand{\CalibrationGapClosedCI}{10.4--38.9\%}
\newcommand{\CalibrationRegretReduction}{0.0115}
\newcommand{\CalibrationRegretReductionCI}{0.0040--0.0194}
\newcommand{\MechanismClaim}{Monotonic recalibration significantly improves routing regret but only partially mediates the pooled-supervision advantage. Pooled supervision also materially improves rollout AUROC, so any routing gain is attributed to overall probabilistic quality rather than calibration alone.}

\newcommand{\reliabilitycomparison}[1]{%
  \begin{figure*}[t]
      \centering

      \begin{minipage}[c]{0.01\linewidth}
          \centering
          \rotatebox{90}{\hspace{+1.9em}\textbf{\texttt{\detokenize{#1}}}}
      \end{minipage}%
      \hfill
      \begin{minipage}[c]{0.97\linewidth}
          \centering

          \begin{subfigure}[t]{0.49\linewidth}
              \centering
              \includegraphics[width=\linewidth]
              {attachments/reliability_english_probe_transfer_grid_#1.pdf}
              \caption{English-trained transfer}
              \label{fig:reliability-#1-english-transfer}
          \end{subfigure}%
          \hfill
          \begin{subfigure}[t]{0.49\linewidth}
              \centering
              \includegraphics[width=\linewidth]
              {attachments/reliability_pooled_multilingual_#1.pdf}
              \caption{Pooled multilingual}
              \label{fig:reliability-#1-pooled}
          \end{subfigure}
      \end{minipage}

      \caption{
      Per-language reliability diagrams for \texttt{\detokenize{#1}} at its
      shared reporting layer. The English-trained probe is transferred unchanged
      to each target language, whereas the pooled probe uses the same total
      training budget distributed equally across the ten languages. The diagonal
      denotes perfect calibration.
      }
      \label{fig:reliability-#1-comparison}
  \end{figure*}%
  }

\section{Results}
\label{sec:results}

Before evaluating probe \textsc{discrimination}, \textsc{calibration}, and \textsc{routing utility}, we first examine the extent to which model success varies across prompt languages. For \texttt{Qwen3-4B}, the empirical success rate ranges from 0.145 in Swahili to 0.716 in English (refer to Table~\ref{tab:qwen3-4b-success-rates}), showing that success depends jointly on the mathematical problem, the evaluated model, and the language in which the problem is presented.
Similar cross-lingual variation is observed across the other evaluated models, with complete model-by-language success rates reported in Table~\ref{tab:base-success-rates}. A useful multilingual probe must therefore recover a model- and language-conditional success signal. 

We subsequently evaluate whether English-trained probes preserve this ordering across languages, whether their predictions remain probabilistically calibrated, and whether they support effective cost-aware routing. Probe-quality results are macro-averaged over the nine non-English target languages, whereas routing is evaluated over all ten languages, including English. Throughout this section, we report \texttt{Qwen3-4B} as the representative model; the corresponding results for all other tested models exhibit consistent trends and are reported in Appendix~\ref{a:additional_results}.

\begin{table}[t!]
    \centering
    \small
    \renewcommand{\arraystretch}{1.15}
    \setlength{\tabcolsep}{12pt}
    \begin{tabular}{lc}
        \toprule
        \textbf{Language}
        & \makecell{\textbf{\texttt{Qwen3-4B}}} \\
        \midrule
        \textit{English} & 0.716 \\
        \textit{Chinese} & 0.630 \\
        \textit{Italian} & 0.621 \\
        \textit{French}  & 0.613 \\
        \textit{Russian} & 0.611 \\
        \textit{Arabic}  & 0.537 \\
        \textit{Thai}    & 0.526 \\
        \textit{Turkish} & 0.526 \\
        \textit{Telugu}  & 0.322 \\
        \textit{Swahili} & 0.145 \\
        \midrule
        \textbf{Mean} & 0.525 \\
        \bottomrule
    \end{tabular}
    \caption{Empirical success rates of \texttt{Qwen3-4B} across languages. Each value is the average empirical success score over the test split, where each problem-level score is computed from five sampled generations.}
    \label{tab:qwen3-4b-success-rates}
\end{table}

\subsection{Cross-Lingual Discrimination}
\label{sec:ranking-results}
\reliabilitycomparison{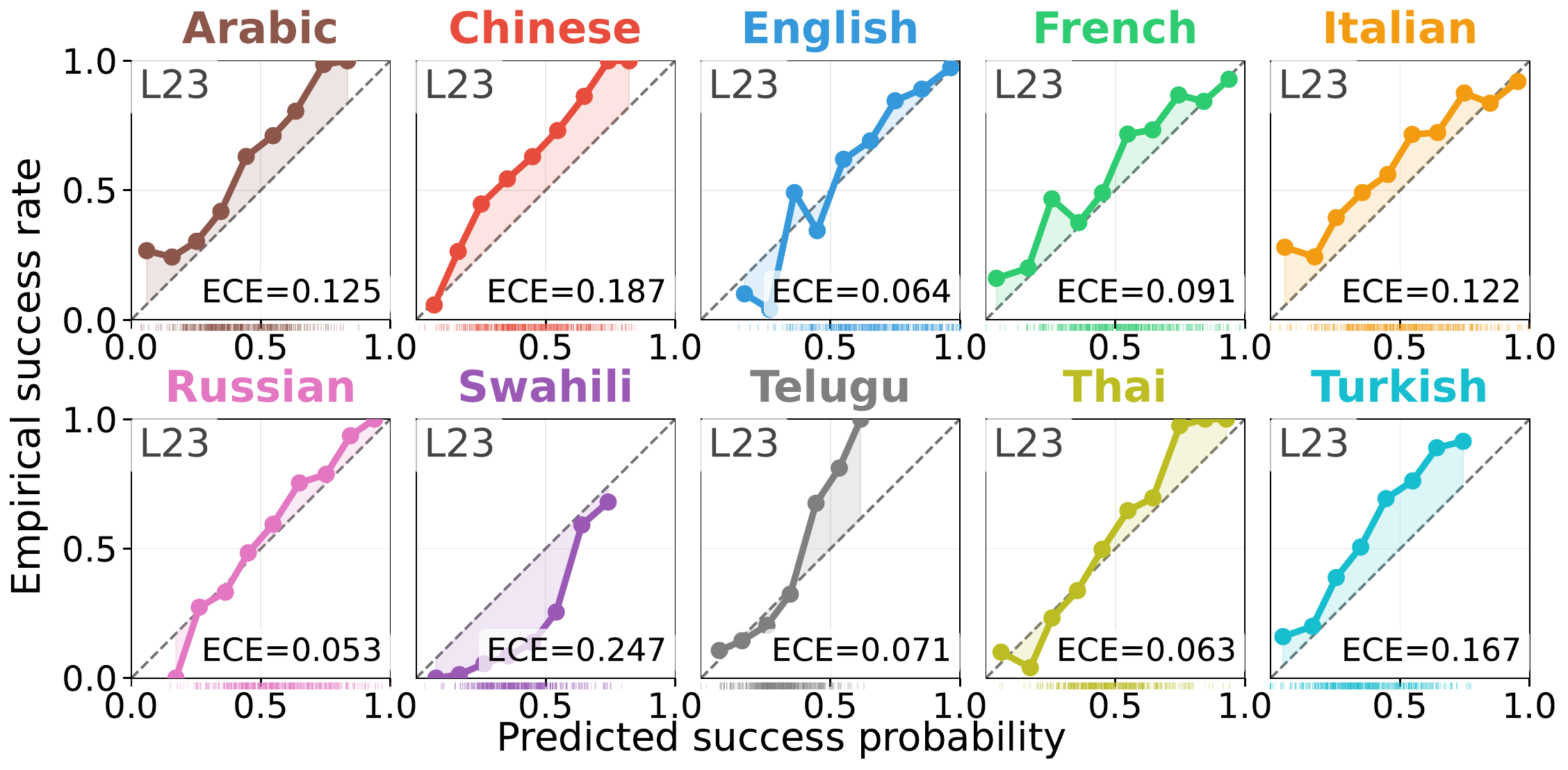}

\definecolor{softred}{RGB}{194,68,68}
\definecolor{softblue}{RGB}{68,105,168}

\begin{table}[b!]
\centering
\small
\setlength{\tabcolsep}{5pt}
\begin{tabular}{lccc}
\toprule
\makecell[l]{\textbf{Evaluation}\\[-1.1pt]\textbf{regime}}
& \makecell{\textbf{Rollout}\\[-1.1pt]\textbf{AUROC $\uparrow$}}
& \makecell{\textbf{Rollout}\\[-1.1pt]\textbf{Brier $\downarrow$}}
& \textbf{ECE $\downarrow$} \\
\midrule
\textit{English}
& 0.729
& 0.213
& 0.125 \\

\textit{Pooled}
& 0.741
& 0.190 
& 0.043  \\

\textit{Oracle}
& 0.758
& 0.183
& 0.031 \\
\bottomrule
\end{tabular}
\caption{Self-probe quality for \texttt{Qwen3-4B} at the shared reporting layer (L23). Values are macro-averages over the nine non-English test languages. AUROC treats the five rollouts as weighted binary outcomes.}
\label{tab:qwen3-4b-self-main}
\end{table}

Table~\ref{tab:qwen3-4b-self-main} shows that the English-trained \texttt{Qwen3-4B} probe retains substantial discrimination when transferred unchanged to the nine non-English languages. It achieves a macro-average rollout AUROC of 0.729, only 2.9 percentage points below the 0.758 obtained by probes trained directly in each evaluation language (\textit{oracle}). Thus, English supervision retains enough success-relevant information to preserve much of the ordering between more and less likely successes after language shift.

Equal-budget pooled multilingual supervision further increases AUROC to 0.741, reducing the gap to the same-language reference from 2.9 to 1.7 percentage points. Because the English and pooled conditions both use 2,100 training prompts, this improvement results from the linguistic composition of the supervision rather than a larger annotation budget. The pattern is consistent across all 8 model configurations (refer to Table~\ref{tab:self-main}). Pooled supervision improves AUROC for every configuration and raises the cross-model mean from 0.697 to 0.724, thereby reducing the average gap to the same-language reference from 4.9 to 2.2 percentage points. Cross-lingual ranking information is therefore broadly transferable from English, although multilingual supervision recovers a substantial portion of the remaining discrimination loss.




\subsection{Ranking Transfers More Reliably Than Probability Scales}
\label{sec:probability-results}
As introduced in Section~\ref{sec:evaluation}, AUROC measures whether a probe orders examples correctly, but not whether its scores match the underlying success probabilities. Table~\ref{tab:qwen3-4b-self-main} shows that English-trained probes retain useful discrimination after transfer, while their Brier scores and ECE deteriorate substantially relative to the labelled same-language oracles. Equal-budget pooled supervision recovers a considerable portion of this loss: it reduces mean Brier score from \MeanSelfEnglishBrier{} to \MeanSelfPooledBrier{} and lowers ECE for all 8 models (refer to Table \ref{tab:self-main}), bringing both metrics closer to the oracle reference. Because Brier score measures overall probabilistic quality, this recovery may reflect improvements in discrimination and calibration; ECE more directly identifies improved alignment between predicted and observed probabilities.  These aggregate metrics do not show whether the calibration loss is uniform across languages or driven by particular language shifts. The per-language reliability diagrams in Figure~\ref{fig:reliability-Qwen3-4B-comparison} therefore examine how the relationship between predicted and observed success changes for each target language in the \texttt{Qwen3-4B} model.

The English-trained probe remains capable of ranking many target-language problems, but the mapping from its scores to observed success rates varies across languages. In particular, the same numerical prediction can correspond to different empirical success frequencies after language shift. Distributing the same supervision budget across languages produces scores that lie closer to the reliability diagonal and are more comparable across the covered language distribution. For reliability diagrams for the other models see Appendix~\ref{a:additional_results}.

\subsection{ETD Success-Prediction Performance}

\label{sec:etd-results}

Same-model probe performance is not sufficient to characterize routing because the ETD router \cite{varshney2026router} does not use each target model's own representations. Instead, to save cost, \texttt{Qwen3-4B} supplies a shared representation, and a separate linear head predicts the success of each candidate target model. We therefore evaluate the ETD heads whose predictions are actually passed to the router. The shared ETD layer is layer \ETDSelectedLayer{}, selected using the joint English--pooled validation-Brier criterion and then frozen for calibration, testing, and routing.

\begin{table}[t]
\centering
\small
\setlength{\tabcolsep}{5pt}
\begin{tabular}{lcccc}
\toprule
\textbf{Predictor}
& \textbf{AUROC $\uparrow$}
& \textbf{Brier $\downarrow$}
& \textbf{NLL $\downarrow$}
& \textbf{ECE $\downarrow$} \\
\midrule
\textit{English raw}
& 0.729
& 0.213
& 0.624
& 0.125 \\

\textit{English + cal.}
& 0.729
& 0.208
& 0.614
& 0.099 \\

\textit{Pooled raw}
& \textbf{0.741}
& \textbf{0.190}
& \textbf{0.573}
& \textbf{0.043} \\
\bottomrule
\end{tabular}
\caption{Prediction quality of the \texttt{Qwen3-4B} ETD routing head at the shared encoder's reporting layer (L23), macro-averaged over the nine non-English test languages. Bold marks the best value for each metric.}
\label{tab:qwen3-4b-etd}
\end{table}

Table~\ref{tab:qwen3-4b-etd} shows that English-trained ETD heads retain some useful ranking information, with mean rollout AUROC \MeanETDEnglishAUROC{}. Nevertheless, pooled supervision improves mean AUROC to \MeanETDPooledAUROC{} and reduces mean Brier score from \MeanETDEnglishBrier{} to \MeanETDPooledBrier{}. The routing advantage of pooled supervision therefore cannot be attributed to calibration alone: pooling improves the overall probabilistic predictions, including their discrimination.

To isolate the contribution of probability scaling, we apply a positive-slope variant of Platt scaling \citep{platt1999probabilistic} to each English-trained target-model head using pooled multilingual validation counts (Appendix~\ref{a:monotonic-recalibration}). This transformation preserves within-model rankings, and hence AUROC, while allowing the probability scale to change. Recalibration reduces mean Brier score from \MeanETDEnglishBrier{} to \MeanETDCalibratedBrier{}, but pooled raw predictions remain better at \MeanETDPooledBrier{}. Calibration metrics generally improve after recalibration, although not uniformly for every target model. This ablation therefore shows that correcting score scales helps, but does not reproduce the full effect of multilingual supervision.

\subsection{Pooled Supervision Produces Stronger Routing Trade-offs}
\label{sec:routing-results}

We finally apply the ETD predictions to the cost-aware policy in Equation~\ref{eq:routing}. We compare the English-trained probe, its monotonically recalibrated variant, the pooled multilingual probe, and the baseline that always selects \texttt{gpt-oss-high}. Each reported operating point is selected on validation and frozen before test evaluation.

\begin{table}[t]
\centering
\scriptsize
\setlength{\tabcolsep}{3pt}
\resizebox{\columnwidth}{!}{%
\begin{tabular}{llrrr}
\toprule
\textbf{Router} & \textbf{Val. choice} & \textbf{$\lambda$} & \textbf{Test success [95\% CI]} & \textbf{Saving [95\% CI]} \\
\midrule
\multirow{2}{*}{\textit{English raw}} & Max. success & 0.00 & 80.8 [78.8,82.8] & 33.7 [31.8,35.5] \\
 & Anchor match & -- & \multicolumn{2}{c}{Not attained on validation} \\
\addlinespace[1pt]
\cmidrule{2-5}
\multirow{2}{*}{\textit{English + cal.}} & Max. success & 0.00 & 85.0 [83.1,86.8] & 10.8 [9.0,12.6] \\
 & Anchor match & -- & \multicolumn{2}{c}{Not attained on validation} \\
\addlinespace[1pt]
\cmidrule{2-5}
\multirow{2}{*}{\textit{Pooled raw}} & Max. success & 0.00 & 86.6 [84.8,88.3] & 13.0 [10.2,15.9] \\
 & Anchor match & 0.10 & 85.5 [83.6,87.2] & 22.9 [19.8,26.3] \\
\addlinespace[1pt]
\cmidrule{2-5}
\textit{Always oss-high} & Baseline & -- & 85.9 & 0.0 \\
\bottomrule
\end{tabular}%
}
\caption{Validation-selected routing operating points. Anchor matching allows at most
0.1 percentage point validation success loss, then minimizes cost; unattained
rows mean no fixed-grid point met that constraint. Brackets are paired bootstrap
intervals over base problem IDs; choices are frozen before test.}
\label{tab:routing-main}
\end{table}

Figure~\ref{fig:routing-frontiers} shows the test-set success--cost frontiers obtained by sweeping the fixed $\lambda$ grid. The figure displays all test outcomes, whereas Table~\ref{tab:routing-main} reports two deployment points whose $\lambda$ values are selected exclusively on validation: maximum success and anchor matching. The selected values are then frozen and evaluated on test.

At its validation-selected maximum-success point, the pooled router obtains 86.6\% test success while saving 13\% of the anchor cost. Its anchor-matching selection saves 22.9\% on test and obtains 85.5\% success. This test success is slightly below the anchor because the matching constraint is enforced on validation; it does not guarantee an exact match after transfer to the test set.


Neither the English probe nor its recalibrated variant meets the anchor-matching criterion on validation. Nevertheless, recalibration increases their validation-selected maximum test success from 80.8\% to 85.0\%. Correcting the probability scales therefore improves routing substantially, but does not fully recover the pooled router's performance.

Across the routing grid, recalibration closes \CalibrationGapClosedPercent{} of the English-to-pooled utility-regret gap (95\% CI: \CalibrationGapClosedCI{}). Calibration thus explains part, but less than half, of the pooled advantage. Because pooling also improves ETD AUROC, its routing benefit comes from both better rankings and more reliable probability estimates.

\section{Conclusion}
\label{sec:conclusion}

This work investigated whether pre-generation success probes can support cost-aware model routing across languages. Using 3,000 mathematical problems in ten languages and 8 open-weight model configurations, we found that English-trained probes retain useful cross-lingual discrimination, meaning they can continue to rank likely successes above failures after language transfer. Their numerical predictions, however, do not preserve the same probabilistic meaning across languages. Equal-budget pooled multilingual supervision improves both discrimination and calibration, producing success estimates that are more reliable across languages and candidate models.
These improvements translate into stronger routing decisions. The pooled multilingual router reaches 86.6\% test success while saving 13.0\% relative to always using the highest-average-success model, and saves 22.9\% at its validation-selected anchor-matching operating point. Monotonic recalibration substantially improves English-trained predictions but recovers only about one quarter of the routing-utility gap, indicating that the benefit of multilingual supervision cannot be explained by calibration alone. Overall, our results demonstrate that transferable rankings are necessary but insufficient for multilingual routing.


\section*{Limitations}
\label{sec:limitations}
Our study is limited to mathematical reasoning, these findings might not extend to other domains. Nine of our ten languages are machine-translated rather than natively authored, so results reflect transfer across translated prompts. We restrict our analysis to linear probes at a single post-prompt position and do not probe during generation, which may have an impact, especially under extended reasoning traces. We evaluate only open-weight models up to 20B parameters, and larger or closed-weight models may behave differently. Finally, our routing costs are estimated under a simplified model rather than measured from a live deployment.





\bibliography{custom}

\appendix
\section{Translation Quality Assessment}
\label{a:translation}

\tcbset{
    promptstyle/.style={
        enhanced,
        breakable,
        colback=gray!4,
        colframe=black!65,
        boxrule=0.5pt,
        arc=1mm,
        left=1.5mm,
        right=1.5mm,
        top=1.5mm,
        bottom=1.5mm,
        fonttitle=\bfseries,
        before skip=0.6\baselineskip,
        after skip=0.6\baselineskip
    }
}

We compare translation quality using the reference-free \texttt{Unbabel/wmt22-cometkiwi-da} model \citep{rei2022cometkiwi}. COMET-Kiwi assigns a relative quality score based on the semantic adequacy of a translation with respect to its English source. The scores are useful for comparing and ranking the two translation systems, but their absolute values do not correspond to an independently interpretable level of translation quality. We score all 3,000 problems in each target language for both DeepL \citep{deepl2026} and Qwen3.5-27B \cite{qwen3.5}. The latter was prompted using the below provided prompt. DeepL obtains a slightly higher mean COMET-Kiwi score than Qwen3.5-27B (see Table \ref{tab:comet-kiwi}), and its translation were therefore selected for our experiments.

\begin{tcolorbox}[promptstyle,title={<System Prompt>}]
You are a professional technical translator specializing in mathematics
competition problems.

\medskip
You will be given a math problem written in English, from the MATH benchmark.
Translate ONLY the natural-language prose into
\textbf{\texttt{[TARGET LANGUAGE]}}. Do not translate, rename, reformat, or
otherwise modify:
\begin{itemize}
    \item Asymptote code blocks, delimited by \texttt{[asy] ... [/asy]};
    \item \LaTeX{} math, delimited by \texttt{\$...\$},
          \texttt{\$\$...\$\$},
          \texttt{\textbackslash{}[...\textbackslash{}]}, or
          \texttt{\textbackslash{}(...\textbackslash{})};
    \item \LaTeX{} commands (e.g.,
          \texttt{\textbackslash{}boxed\{\}},
          \texttt{\textbackslash{}frac\{\}\{\}},
          \texttt{\textbackslash{}sqrt\{\}},
          \texttt{\textbackslash{}left},
          \texttt{\textbackslash{}right}, and
          \texttt{\textbackslash{}text\{\}});
    \item variable names, numbers, units, and mathematical symbols that
          appear inside formulas.
\end{itemize}
Preserve the original paragraph structure and line breaks. Do not solve the
problem. Do not add commentary, notes, or explanations of any kind---translate
what is given, nothing more.

\medskip
Respond with a single JSON object of the form
\texttt{\{"translation": "<translated text>"\}} and nothing else.
\end{tcolorbox}

\begin{tcolorbox}[promptstyle,title={<User Prompt>}]
Translate the following math problem into
\textbf{\texttt{[TARGET LANGUAGE]}}.

\medskip
\texttt{<<PROBLEM>>}

\textbf{\texttt{[SOURCE PROBLEM]}}

\texttt{<<END>>}
\end{tcolorbox}

\begin{table}[h!]
\centering
\small
\begin{tabular}{lcc}
\toprule
\textbf{Language} & \textbf{DeepL} & \textbf{Qwen3.5-27B} \\
\midrule
\textit{Arabic}  & \textbf{0.7758} & 0.7677 \\
\textit{Chinese} & 0.8050 & \textbf{0.8058} \\
\textit{French}  & \textbf{0.8293} & 0.8286 \\
\textit{Italian} & \textbf{0.8295} & 0.8278 \\
\textit{Russian} & \textbf{0.8139} & 0.8121 \\
\textit{Swahili} & \textbf{0.7913} & 0.7834 \\
\textit{Telugu}  & \textbf{0.7934} & 0.7826 \\
\textit{Thai}    & 0.7921 & \textbf{0.7942} \\
\textit{Turkish} & \textbf{0.8284} & 0.7940 \\
\midrule
\textbf{Overall} & \textbf{0.8065} & 0.7996 \\
\bottomrule
\end{tabular}
\caption{Mean COMET-Kiwi scores over 3,000 MATH problems per target language
and translation system. Higher is better; bold indicates the stronger system
within each row.}
\label{tab:comet-kiwi}
\end{table}

\section{Layer-wise Ridge Regularization Selection}
\label{app:alpha-selection}

\begin{figure}[t]
  \centering
  \includegraphics[width=\linewidth]{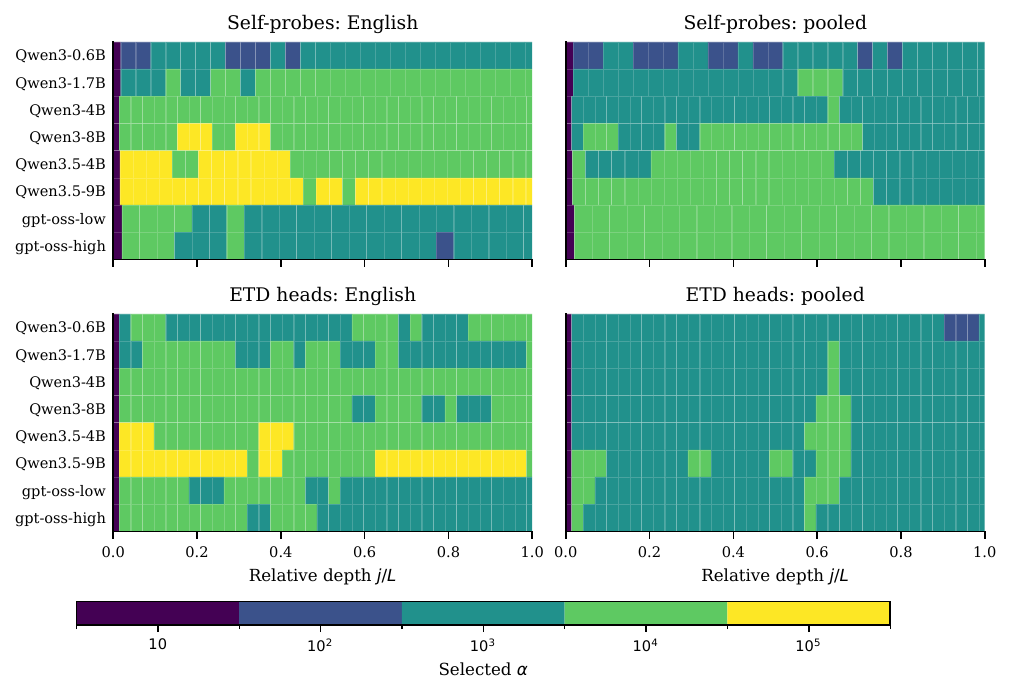}
  \caption{Ridge coefficient selected at every relative representation depth.
  The top row shows the same-model probe and the bottom row shows ETD target-specific
  heads; columns separate English and pooled supervision. In each cell, the
  selected $\alpha$ minimizes rollout-level validation Brier score over
  $\{10,10^2,10^3,10^4,10^5\}$. Each row denotes the model supplying the
  representation for the same-model probe and the predicted target model for ETD; each
  cell denotes one layer. ETD heads use the shared \texttt{Qwen3-4B}
  representation.}
  \label{fig:alpha-selection}
\end{figure}

For every model, representation layer, and supervision regime, we fit one
ridge probe for each
$\alpha\in\{10,10^2,10^3,10^4,10^5\}$. Predictions are projected onto
$[0,1]$, and the coefficient with the lowest rollout-level Brier score on the
corresponding validation set is selected. English probes use English
validation examples, whereas pooled probes use the combined, equally sized
validation sets from all ten languages. ETD heads apply the same rule
separately to each target model at every layer of the shared
\texttt{Qwen3-4B} encoder. Exact validation-Brier ties are resolved in favour
of the smaller value of $\alpha$. Figure~\ref{fig:alpha-selection} focuses on
the English and pooled regimes used for shared-layer selection and routing;
the same-language probes use the identical selection rule.

\section{Why Discrimination, Calibration, and Probabilistic Quality Differ}
\label{a:metric-differences}

Consider a balanced dataset containing two equally large groups of problems.
Generations for the ``easy'' group succeed with probability $0.75$, while
generations for the ``hard'' group succeed with probability $0.25$. The
overall success rate is therefore $0.5$. Table~\ref{tab:metric-example}
compares three hypothetical probes.

\begin{table}[h]
  \centering
  \tiny
  \setlength{\tabcolsep}{4pt}
  \begin{tabular}{lrrrrrr}
      \toprule
      \textbf{Probe}
      & $\widehat{p}_{\mathrm{easy}}$
      & $\widehat{p}_{\mathrm{hard}}$
      & \textbf{AUROC}
      & \textbf{ECE}
      & \textbf{Brier}
      & \textbf{NLL} \\
      \midrule
      \textit{Constant}       & $0.50$ & $0.50$ & $0.50$ & $0.00$ & $0.250$ & $0.693$ \\
      \textit{Overconfident}  & $1.00$ & $0.00$ & $0.75$ & $0.25$ & $0.250$ & $\infty$ \\
      \textit{Well specified} & $0.75$ & $0.25$ & $0.75$ & $0.00$ & $0.188$ & $0.562$ \\
      \bottomrule
  \end{tabular}
  \caption{
      Expected metrics for three hypothetical probes. The dataset contains
      equally many easy and hard problems with true success probabilities
      $0.75$ and $0.25$, respectively. The infinite NLL for the
      overconfident probe is the theoretical value before numerical clipping.
  }
  \label{tab:metric-example}
\end{table}

The overconfident and well-specified probes have identical AUROC because they produce the same ordering: both assign larger scores to easy problems than to hard problems. AUROC cannot reveal that the first probe's probabilities are too extreme. ECE distinguishes them by comparing their numerical predictions with the observed group frequencies.

Conversely, the constant and well-specified probes both have zero ECE. The constant probe predicts the correct overall prevalence and is therefore calibrated in aggregate, but it cannot distinguish easy from hard problems and has AUROC $0.5$. Thus, calibration alone does not establish that a predictor is informative.

Finally, the constant and overconfident probes have the same expected Brier score despite failing for different reasons. The constant probe is calibrated but uninformative, whereas the overconfident probe is discriminative but miscalibrated. AUROC and ECE reveal this distinction. NLL additionally exposes the severity of the overconfident errors: assigning probability zero to an event that occurs produces an infinite logarithmic penalty. In practice, we clip probabilities only for numerical evaluation, resulting in a finite but still large penalty.


\section{Routing Cost Estimation}
\label{a:routing-cost}

\begin{table*}[t]
\centering
\small
\renewcommand{\arraystretch}{1.15}
\begin{tabular}{lcccccc}
\toprule
\textbf{Model} &
\textbf{Input price} &
\textbf{Output price} &
\textbf{Avg. input} &
\textbf{Avg. output} &
\textbf{Exp. cost} &
\textbf{Relative} \\
&
\textbf{(\$/M tok.)} &
\textbf{(\$/M tok.)} &
\textbf{tokens} &
\textbf{tokens} &
\textbf{(\$/k gen.)} &
\textbf{Exp. cost}
\\
\midrule
\texttt{\textbf{Qwen3-0.6B}}   & 0.010 & 0.060 & 205.12 & 593.45  & 0.0377 & 0.0111 \\
\texttt{\textbf{Qwen3-1.7B}}   & 0.028 & 0.170 & 205.12 & 2361.62 & 0.4073 & 0.1196 \\
\texttt{\textbf{Qwen3-4B}}     & 0.067 & 0.400 & 205.12 & 2056.09 & 0.8361 & 0.2456 \\
\texttt{\textbf{gpt-oss-low}}  & 0.204 & 1.224 & 226.57 & 946.03  & 1.2042 & 0.3537 \\
\texttt{\textbf{Qwen3.5-4B}}   & 0.067 & 0.400 & 186.49 & 3042.49 & 1.2294 & 0.3611 \\
\texttt{\textbf{Qwen3-8B}}     & 0.133 & 0.800 & 205.12 & 1545.65 & 1.2639 & 0.3712 \\
\texttt{\textbf{Qwen3.5-9B}}   & 0.150 & 0.900 & 184.49 & 2538.44 & 2.3123 & 0.6791 \\
\texttt{\textbf{gpt-oss-high}} & 0.204 & 1.224 & 226.57 & 2743.93 & 3.4048 & 1.0000 \\
\bottomrule
\end{tabular}
\caption{Proxy prices used in the routing simulation. Average input lengths are computed over the balanced pooled training sample used for routing-cost estimation (210 prompts per language and 2,100 prompts per model), using each model's native tokenizer and chat template. Average output lengths are computed over the same sample. Relative expected costs are normalized by the average standalone input-plus-output cost of \texttt{gpt-oss-high}.}
\label{tab:routing-model-costs}
\end{table*}

Because the evaluated open-weight models do not have directly comparable first-party API prices, we construct a deployment-aware proxy for their inference costs. The proxy prices reflect each model's memory footprint under the precision used for inference, rather than its nominal parameter count alone.

Let $M_m$ denote the deployed memory footprint of model $m$, in GB. We assign the output-token price according to
\begin{equation*}
P_m^{\mathrm{out}}
=
\alpha M_m,
\qquad
\alpha = 0.10,
\end{equation*}
where $P_m^{\mathrm{out}}$ is expressed in US dollars per million output tokens. This proxy is intended to preserve relative deployment costs across models rather than reproduce the prices of a particular inference provider. The deployed memory footprint is particularly relevant for \texttt{gpt-oss-20b}, whose weights are natively distributed in a quantized
format.\footnote{%
  {\fontsize{8.5pt}{9pt}\selectfont
  \url{https://openai.com/index/introducing-gpt-oss/}}%
}Its price is therefore based on its quantized inference footprint rather than its nominal parameter count, resulting in a lower estimated cost than would be assigned to an unquantized 20-billion-parameter model. For every model, the input-token price is set to one sixth of its output-token price:
\begin{equation*}
P_m^{\mathrm{in}}
=
\frac{1}{6}P_m^{\mathrm{out}}.
\end{equation*}
This ratio is similar to the input-to-output pricing ratio used by several
hosted closed-source models.\footnote{%
  {\fontsize{8.5pt}{9pt}\selectfont
  \url{https://developers.openai.com/api/docs/pricing}}%
}

The \texttt{gpt-oss-low} and \texttt{gpt-oss-high} configurations are assigned
the same input and output prices because they use the same model weights and
have the same memory footprint. Their only difference is the configured
reasoning effort. The resulting difference in computational cost is therefore
captured through their observed input and output token counts: the
high-reasoning configuration processes and generates more tokens on average,
despite having the same per-token price.

For each model, let $P_m^{\mathrm{in}}$ and $P_m^{\mathrm{out}}$ denote the
input and output prices reported in Table~\ref{tab:routing-model-costs}. Let
$T_{m,i}^{\mathrm{in}}$ and $T_{m,i}^{\mathrm{out}}$ denote the corresponding
input and output token counts when model $m$ processes example $i$. The
estimated generation cost is
\begin{equation*}
c_{m,i}
=
\frac{
P_m^{\mathrm{in}}T_{m,i}^{\mathrm{in}}
+
P_m^{\mathrm{out}}T_{m,i}^{\mathrm{out}}
}{10^6}.
\end{equation*}
Thus, the cost difference between \texttt{gpt-oss-low} and \texttt{gpt-oss-high} arises from their different observed token counts rather than from different per-token prices.

For ETD routing, each query is first processed by the shared \texttt{Qwen3-4B} encoder before a target model is selected. The total cost of routing example $i$ to model $m$ therefore additionally includes the input-prefilling cost of this encoder:
\begin{equation*}
    c^{\mathrm{ETD}}_{m,i}
  =
  \begin{cases}
  c_{m,i},
  & m = \texttt{Qwen3-4B}, \\[3pt]
  c_i^{\mathrm{encoder}} + c_{m,i},
  & m \neq \texttt{Qwen3-4B}.
  \end{cases}
\end{equation*}
where
\begin{equation*}
c_i^{\mathrm{encoder}}
=
\frac{
P_{\mathrm{Qwen3\text{-}4B}}^{\mathrm{in}}
T_{\mathrm{Qwen3\text{-}4B},i}^{\mathrm{in}}
}{10^6}.
\end{equation*}
Because the shared encoder is used only to extract the internal representation required by the routing policy, it does not generate output tokens and therefore incurs no output-token cost. When the selected target is \texttt{Qwen3-4B}, its encoder prefill is reused for generation rather than charged a second time.

Expected costs are obtained by averaging the per-example costs after applying the routing policy. For a policy $\pi$ that selects model $\pi(i)$ for example $i$, the expected ETD routing cost over an evaluation set containing $N$ examples is
\begin{equation*}
C^{\mathrm{ETD}}(\pi)
=
\frac{1}{N}
\sum_{i=1}^{N}
c^{\mathrm{ETD}}_{\pi(i),i}.
\end{equation*}

Relative costs are reported with respect to always selecting
\texttt{gpt-oss-high}:
\begin{equation*}
C^{\mathrm{rel}}(\pi)
=
\frac{
C^{\mathrm{ETD}}(\pi)
}{
\frac{1}{N}
\sum_{i=1}^{N}
c_{\mathrm{gpt\text{-}oss\text{-}high},i}
}.
\end{equation*}
The reference-system denominator excludes the shared encoder cost because the always-\texttt{gpt-oss-high} policy invokes the target model directly and does not perform an ETD routing step.

\section{Shared Encoder Selection}
\label{a:encoder-selection}

\paragraph{Search procedure.}
We selected the shared encoder using a validation-only diagnostic conducted
before the final probe-selection pipeline was fixed. Starting from the
same 8 model configurations considered as routing targets, we treat each
configuration in turn as the ETD encoder. For every encoder, we train pooled
multilingual target-specific ridge heads at each available representation
layer. We then evaluate the resulting router on the common validation split,
using the same candidate set, cost model, routing policies, $\lambda$ values,
and anchor-margin grid for every encoder.

This diagnostic differs from the final probe-training protocol in one respect
that matters for reproducibility. Its ridge grid is
$\alpha\in\{10,10^2,10^3,10^4,10^5\}$, and it selects $\alpha$
separately for every encoder layer and target head by maximum validation AUROC,
breaking ties by validation RMSE and then smaller $\alpha$. The main pipeline,
by contrast, selects ridge coefficients by validation Brier score as described
in Section~\ref{sec:probes}. We therefore use the all-encoder scan only
to motivate the encoder architecture; all reported \texttt{Qwen3-4B} probe,
calibration, and routing results are retrained under the final protocol.

Because the architectures have different numbers of layers, comparing all
encoders at one absolute layer would be arbitrary. We therefore first select,
separately for each encoder, the nonzero layer whose pooled validation routing
frontier attains the highest success; layer 0 is evaluated but excluded from
this maximum-success selection. Table~\ref{tab:encoder-choice} reports that layer
and the maximum validation success attained on its frontier. This criterion is
used only for the encoder comparison. After fixing \texttt{Qwen3-4B}, the main
pipeline selects the reporting layer using the joint validation-Brier rule in
Equation~\ref{eq:shared-layer}; that separate rule also selects layer 23. The
encoder and reporting layer are frozen before any test evaluation.

\begin{table}[t!]
\centering
\scriptsize
\setlength{\tabcolsep}{3.5pt}
\begin{tabular}{lrrr}
\toprule
Shared encoder & Layer & \makecell{Max. validation\\success} & \makecell{Prefill /\\anchor} \\
\midrule
\texttt{Qwen3-0.6B}         & 21 & 88.1\% & 0.06\% \\
\texttt{Qwen3-1.7B}         & 19 & 87.9\% & 0.17\% \\
\textbf{\texttt{Qwen3-4B}} & \textbf{23} & \textbf{88.3\%} & \textbf{0.40\%} \\
\texttt{Qwen3-8B}           & 26 & 88.1\% & 0.80\% \\
\texttt{Qwen3.5-4B}         & 10 & 88.2\% & 0.37\% \\
\texttt{Qwen3.5-9B}         & 13 & 88.2\% & 0.81\% \\
\texttt{gpt-oss-high}       &  6 & 88.1\% & 1.36\% \\
\texttt{gpt-oss-low}        &  6 & 87.9\% & 1.36\% \\
\bottomrule
\end{tabular}
\caption{Validation-only shared-encoder comparison under pooled ETD
supervision. For each encoder, the table reports its validation-selected layer
and the maximum success attained over the fixed validation routing grid. No
test outcomes enter the comparison. Prefill cost is the encoder's standalone
input-only cost as a percentage of the standalone \texttt{gpt-oss-high} anchor
cost, using the prices and training-mean input lengths in
Appendix~\ref{a:routing-cost}. }
\label{tab:encoder-choice}
\end{table}

\paragraph{Interpretation.}
\texttt{Qwen3-4B} obtains the highest observed maximum validation success,
88.3\%, with a nominal prefill cost equal to 0.40\% of the anchor cost. The
comparison is close: \texttt{Qwen3.5-4B} obtains 88.2\% at a slightly lower
prefill cost of 0.37\%, while \texttt{Qwen3.5-9B} also obtains 88.2\% but has
roughly twice the prefill cost of \texttt{Qwen3-4B}. Smaller Qwen3 encoders are
cheaper, but have lower maximum validation success in this experiment. We
therefore choose \texttt{Qwen3-4B} as the observed validation leader with a
moderate prefill requirement. Since the differences among the leading
encoders are small and the comparison has no uncertainty intervals, it should
not be interpreted as showing that \texttt{Qwen3-4B} is intrinsically better
than the alternatives.

The prefill column provides a common cost reference rather than the realized
overhead of every route. When the router selects \texttt{Qwen3-4B} itself as
the target model, its encoder prefill is reused for generation and is not
charged a second time.

\begin{table*}[h!]
    \centering
    \scriptsize
    \renewcommand{\arraystretch}{1.15}
    \setlength{\tabcolsep}{4pt}
        \begin{tabular}{lcccccccc|c}
            \toprule
            \textbf{Language}
            & \makecell{\textbf{\texttt{Qwen3-0.6B}}}
            & \makecell{\textbf{\texttt{Qwen3-1.7B}}}
            & \makecell{\textbf{\texttt{Qwen3-4B}}}
            & \makecell{\textbf{\texttt{Qwen3-8B}}}
            & \makecell{\textbf{\texttt{Qwen3.5-4B}}}
            & \makecell{\textbf{\texttt{Qwen3.5-9B}}}
            & \makecell{\textbf{\texttt{gpt-oss-low}}}
            & \makecell{\textbf{\texttt{gpt-oss-high}}}
            & \textbf{Mean} \\
            \midrule
            \textit{English} & 0.279 & 0.508 & 0.716 & 0.728 & 0.916 & 0.935 & 0.895 & 0.925 & 0.738 \\
            \textit{Chinese} & 0.220 & 0.408 & 0.630 & 0.638 & 0.851 & 0.898 & 0.754 & 0.886 & 0.661 \\
            \textit{French}  & 0.179 & 0.388 & 0.613 & 0.663 & 0.848 & 0.895 & 0.789 & 0.911 & 0.661 \\
            \textit{Russian} & 0.170 & 0.384 & 0.611 & 0.661 & 0.837 & 0.889 & 0.761 & 0.884 & 0.650 \\
            \textit{Italian} & 0.160 & 0.378 & 0.621 & 0.654 & 0.835 & 0.880 & 0.755 & 0.897 & 0.648 \\
            \textit{Arabic}  & 0.134 & 0.312 & 0.537 & 0.600 & 0.785 & 0.841 & 0.749 & 0.885 & 0.605 \\
            \textit{Turkish} & 0.107 & 0.282 & 0.526 & 0.574 & 0.760 & 0.821 & 0.754 & 0.882 & 0.588 \\
            \textit{Thai}    & 0.138 & 0.324 & 0.526 & 0.578 & 0.590 & 0.715 & 0.695 & 0.826 & 0.549 \\
            \textit{Telugu}  & 0.056 & 0.153 & 0.322 & 0.435 & 0.497 & 0.580 & 0.716 & 0.849 & 0.451 \\
            \textit{Swahili} & 0.034 & 0.124 & 0.145 & 0.418 & 0.536 & 0.683 & 0.587 & 0.647 & 0.397 \\
            \midrule
            \textbf{Mean} & 0.148 & 0.326 & 0.525 & 0.595 & 0.746 & 0.814 & 0.746 & 0.859 & 0.595 \\
            \bottomrule
        \end{tabular}
    \caption{Empirical success rates of the evaluated models across languages. Each value is the average empirical success score over the test split, where each problem-level score is computed from five sampled generations. Languages are ordered by decreasing average success rate across models. The final row reports the average success rate for each model across languages, and the final column reports the average success rate for each language across models.}
    \label{tab:base-success-rates}
\end{table*}

\definecolor{softred}{RGB}{194,68,68}
\definecolor{softblue}{RGB}{68,105,168}

\begin{table*}[h!]
\centering
\scriptsize
\setlength{\tabcolsep}{3.5pt}
\begin{tabular}{lcccccccccc}
\toprule
& & \multicolumn{3}{c}{\textbf{Rollout AUROC $\uparrow$}} &
\multicolumn{3}{c}{\textbf{Rollout Brier $\downarrow$}} &
\multicolumn{3}{c}{\textbf{ECE $\downarrow$}} \\
\cmidrule(lr){3-5}\cmidrule(lr){6-8}\cmidrule(lr){9-11}
\textbf{Model} & \textbf{Layer}
& \makecell{\textbf{Eng.}\\$(\Delta_{\mathrm{pp}})$}
& \makecell{\textbf{Pool}\\$(\Delta_{\mathrm{pp}})$}
& \textbf{Oracle}
& \makecell{\textbf{Eng.}\\$(\Delta_{\mathrm{rel}})$}
& \makecell{\textbf{Pool}\\$(\Delta_{\mathrm{rel}})$}
& \textbf{Oracle}
& \makecell{\textbf{Eng.}\\$(\Delta_{\mathrm{rel}})$}
& \makecell{\textbf{Pool}\\$(\Delta_{\mathrm{rel}})$}
& \textbf{Oracle} \\
\midrule

\texttt{\textbf{Qwen3-0.6B}} & 14
& 0.710 (\textcolor{softred}{$-5.9$})
& 0.718 (\textcolor{softred}{$-5.1$})
& 0.769
& 0.120 (\textcolor{softred}{$+26\%$})
& 0.099 (\textcolor{softred}{$+4\%$})
& 0.095
& 0.116 (\textcolor{softred}{$+274\%$})
& 0.038 (\textcolor{softred}{$+23\%$})
& 0.031 \\

\texttt{\textbf{Qwen3-1.7B}} & 20
& 0.741 (\textcolor{softred}{$-3.7$})
& 0.766 (\textcolor{softred}{$-1.2$})
& 0.778
& 0.190 (\textcolor{softred}{$+21\%$})
& 0.161 (\textcolor{softred}{$+3\%$})
& 0.157
& 0.129 (\textcolor{softred}{$+249\%$})
& 0.042 (\textcolor{softred}{$+14\%$})
& 0.037 \\

\texttt{\textbf{Qwen3-4B}} & 23
& 0.729 (\textcolor{softred}{$-2.9$})
& 0.741 (\textcolor{softred}{$-1.7$})
& 0.758
& 0.213 (\textcolor{softred}{$+16\%$})
& 0.190 (\textcolor{softred}{$+4\%$})
& 0.183
& 0.125 (\textcolor{softred}{$+303\%$})
& 0.043 (\textcolor{softred}{$+39\%$})
& 0.031 \\

\texttt{\textbf{Qwen3-8B}} & 21
& 0.739 (\textcolor{softred}{$-1.6$})
& 0.743 (\textcolor{softred}{$-1.2$})
& 0.755
& 0.212 (\textcolor{softred}{$+10\%$})
& 0.196 (\textcolor{softred}{$+2\%$})
& 0.192
& 0.118 (\textcolor{softred}{$+237\%$})
& 0.037 (\textcolor{softred}{$+6\%$})
& 0.035 \\

\texttt{\textbf{Qwen3.5-4B}} & 19
& 0.684 (\textcolor{softred}{$-7.2$})
& 0.737 (\textcolor{softred}{$-1.9$})
& 0.756
& 0.178 (\textcolor{softred}{$+19\%$})
& 0.155 (\textcolor{softred}{$+3\%$})
& 0.150
& 0.083 (\textcolor{softred}{$+186\%$})
& 0.042 (\textcolor{softred}{$+45\%$})
& 0.029 \\

\texttt{\textbf{Qwen3.5-9B}} & 14
& 0.690 (\textcolor{softred}{$-4.9$})
& 0.723 (\textcolor{softred}{$-1.6$})
& 0.739
& 0.148 (\textcolor{softred}{$+15\%$})
& 0.133 (\textcolor{softred}{$+3\%$})
& 0.129
& 0.060 (\textcolor{softred}{$+107\%$})
& 0.036 (\textcolor{softred}{$+24\%$})
& 0.029 \\

\texttt{\textbf{gpt-oss-low}} & 17
& 0.681 (\textcolor{softred}{$-4.0$})
& 0.695 (\textcolor{softred}{$-2.6$})
& 0.721
& 0.183 (\textcolor{softred}{$+6\%$})
& 0.178 (\textcolor{softred}{$+3\%$})
& 0.172
& 0.053 (\textcolor{softred}{$+83\%$})
& 0.033 (\textcolor{softred}{$+14\%$})
& 0.029 \\

\texttt{\textbf{gpt-oss-high}} & 7
& 0.603 (\textcolor{softred}{$-8.7$})
& 0.669 (\textcolor{softred}{$-2.1$})
& 0.690
& 0.125 (\textcolor{softred}{$+12\%$})
& 0.116 (\textcolor{softred}{$+4\%$})
& 0.112
& 0.078 (\textcolor{softred}{$+136\%$})
& 0.030 (\textcolor{softblue}{$-9\%$})
& 0.033 \\

\midrule
\textbf{Mean} & --
& 0.697 (\textcolor{softred}{$-4.9$})
& 0.724 (\textcolor{softred}{$-2.2$})
& 0.746
& 0.171 (\textcolor{softred}{$+15\%$})
& 0.154 (\textcolor{softred}{$+3\%$})
& 0.149
& 0.095 (\textcolor{softred}{$+197\%$})
& 0.038 (\textcolor{softred}{$+19\%$})
& 0.032 \\

\bottomrule
\end{tabular}

\caption{
Self-probe quality at the shared reporting layer. Values are macro-averages
over the nine non-English test languages. Eng. uses 2,100 English prompts,
Pool uses 210 prompts per language (2,100 total), and Oracle uses 2,100
prompts in each target language. Parentheses report deltas relative to
Oracle. For AUROC,
$\Delta_{\mathrm{pp}}=100(X-\mathrm{Oracle})$ is the absolute difference in
percentage points. For Brier and ECE,
$\Delta_{\mathrm{rel}}=100(X-\mathrm{Oracle})/\mathrm{Oracle}$ is the relative
percentage difference. Red indicates worse performance than Oracle and blue
indicates better performance. AUROC treats the five rollouts as weighted
binary outcomes.
}
\label{tab:self-main}
\end{table*}

\section{Additional Results}
\label{a:additional_results}

\subsection{Empirical Success Rates}

Table~\ref{tab:base-success-rates}  reports the empirical success rates of the
evaluated models across all ten languages. Each value is obtained by averaging
the empirical success scores over the test split. Since each problem-level score
is estimated from five sampled generations, these values estimate the
probability that a model solves a problem under the decoding procedure used in
this work, rather than the outcome of a single deterministic generation.

Success varies substantially across both models and languages. English has the
highest average success rate across models, followed by Chinese and French,
whereas Swahili and Telugu have the lowest averages. Increasing model capacity
or reasoning effort generally improves success: performance rises across the
\texttt{Qwen3} model sizes, from \texttt{Qwen3.5-4B} to
\texttt{Qwen3.5-9B}, and from \texttt{gpt-oss-low} to
\texttt{gpt-oss-high}. Nevertheless, the language gaps remain visible even for
the strongest configurations. For example, \texttt{gpt-oss-high} decreases
from 0.925 in English to 0.647 in Swahili. These results reinforce that success
is conditional not only on the underlying problem and target model, but also on
the language in which the problem is presented.

\subsection{Same-model Probe Discrimination and Calibration}
\label{a}

The complete same-model probe results (Table \ref{tab:self-main}) show that the main findings are consistent
across model configurations. English-trained probes retain useful
discrimination after transfer to the nine non-English languages, obtaining a
mean rollout AUROC of \MeanSelfEnglishAUROC{}. Thus, the representations learned
from English supervision preserve a substantial amount of information about
which target-language generations are likely to succeed.

Equal-budget pooled multilingual supervision improves mean AUROC to
\MeanSelfPooledAUROC{}, bringing discrimination closer to the labelled
same-language reference. The improvement indicates that the pooled probe does
not merely adjust the numerical scale of the predictions: multilingual
supervision also improves the ordering of successful and unsuccessful
generations.

The difference between supervision regimes is clearer for probabilistic
quality. English transfer obtains a mean Brier score of
\MeanSelfEnglishBrier{}, whereas pooled supervision reduces it to
\MeanSelfPooledBrier{}. Pooled supervision also lowers ECE for all 8
models, moving the predictions toward the same-language reference. English
supervision therefore transfers more reliably as a ranking function than as a
probability estimator. A score learned in English may still identify relatively
easy and difficult target-language problems, while no longer corresponding to
the same empirical success frequency after language shift.

The same-language probes generally remain the strongest reference because they
are trained with the full labelled training set from the evaluation language.
Importantly, however, pooled supervision uses the same total number of training
examples as the English-only condition. Its advantage therefore results from
the linguistic composition of the supervision rather than from a larger
annotation budget.

\subsection{Encoder--Target Decoupled Prediction Quality}
\label{a}

\begin{table}[t!]
\centering
\scriptsize
\setlength{\tabcolsep}{3pt}
\begin{tabular}{llrrrr}
\toprule
\textbf{Target} & \textbf{Predictor} & \textbf{AUROC $\uparrow$} & \textbf{Brier $\downarrow$} & \textbf{NLL $\downarrow$} & \textbf{ECE $\downarrow$} \\
\midrule
\multirow{3}{*}{\texttt{\textbf{Qwen3-0.6B}}} & English raw & 0.769 & 0.109 & 0.465 & 0.112 \\
 & English + cal. & 0.769 & 0.106 & \textbf{0.385} & 0.093 \\
 & Pooled raw & \textbf{0.773} & \textbf{0.088} & 0.400 & \textbf{0.035} \\
\addlinespace[1pt] \cmidrule{2-6}
\multirow{3}{*}{\texttt{\textbf{Qwen3-1.7B}}} & English raw & 0.759 & 0.176 & 0.547 & 0.103 \\
 & English + cal. & 0.759 & 0.172 & 0.539 & 0.085 \\
 & Pooled raw & \textbf{0.777} & \textbf{0.158} & \textbf{0.520} & \textbf{0.051} \\
\addlinespace[1pt] \cmidrule{2-6}
\multirow{3}{*}{\texttt{\textbf{Qwen3-4B}}} & English raw & 0.729 & 0.213 & 0.624 & 0.125 \\
 & English + cal. & 0.729 & 0.208 & 0.614 & 0.099 \\
 & Pooled raw & \textbf{0.741} & \textbf{0.190} & \textbf{0.573} & \textbf{0.043} \\
\addlinespace[1pt] \cmidrule{2-6}
\multirow{3}{*}{\texttt{\textbf{Qwen3-8B}}} & English raw & 0.723 & 0.226 & 0.648 & 0.138 \\
 & English + cal. & 0.723 & 0.208 & 0.609 & 0.074 \\
 & Pooled raw & \textbf{0.733} & \textbf{0.200} & \textbf{0.606} & \textbf{0.038} \\
\addlinespace[1pt] \cmidrule{2-6}
\multirow{3}{*}{\texttt{\textbf{Qwen3.5-4B}}} & English raw & 0.692 & 0.193 & 0.571 & 0.133 \\
 & English + cal. & 0.692 & 0.190 & 0.566 & 0.141 \\
 & Pooled raw & \textbf{0.727} & \textbf{0.158} & \textbf{0.492} & \textbf{0.040} \\
\addlinespace[1pt] \cmidrule{2-6}
\multirow{3}{*}{\texttt{\textbf{Qwen3.5-9B}}} & English raw & 0.668 & 0.161 & 0.503 & 0.098 \\
 & English + cal. & 0.668 & 0.146 & 0.455 & 0.061 \\
 & Pooled raw & \textbf{0.715} & \textbf{0.134} & \textbf{0.426} & \textbf{0.032} \\
\addlinespace[1pt] \cmidrule{2-6}
\multirow{3}{*}{\texttt{\textbf{gpt-oss-low}}} & English raw & 0.584 & 0.232 & 2.110 & 0.189 \\
 & English + cal. & 0.584 & 0.196 & 0.581 & 0.061 \\
 & Pooled raw & \textbf{0.712} & \textbf{0.174} & \textbf{0.522} & \textbf{0.032} \\
\addlinespace[1pt] \cmidrule{2-6}
\multirow{3}{*}{\texttt{\textbf{gpt-oss-high}}} & English raw & 0.545 & 0.140 & 1.350 & 0.123 \\
 & English + cal. & 0.545 & 0.126 & \textbf{0.419} & 0.057 \\
 & Pooled raw & \textbf{0.693} & \textbf{0.114} & 0.427 & \textbf{0.048} \\
  \midrule
  \multicolumn{6}{l}{\textit{Shared encoder for all rows: \texttt{Qwen3-4B}, layer 23.}} \\
  \multicolumn{6}{l}{\textit{A separate ridge head predicts each target model's success.}} \\
  \bottomrule
\end{tabular}
\caption{Prediction quality of the ETD routing heads, macro-averaged over the nine non-English test languages. English calibration is fit on multilingual validation data with a positive slope. Bold marks the best value per metric within each target model's three predictor settings.}
\label{tab:etd-main}
\end{table}

The ETD results evaluate the predictions that are directly consumed by the
router. Unlike same-model probes, these heads predict the success of every target
model using a shared representation extracted from \texttt{Qwen3-4B}. This is a
more demanding setting because the encoder representation must contain
information not only about its own expected success, but also about the
behaviour of several distinct target models.

English-trained ETD heads retain some discriminative information, with mean
rollout AUROC \MeanETDEnglishAUROC{}. Pooled multilingual supervision improves
this value to \MeanETDPooledAUROC{} and reduces mean Brier score from
\MeanETDEnglishBrier{} to \MeanETDPooledBrier{}. The improvement in AUROC shows
that the pooled routing advantage cannot be explained solely by better
calibration: pooled supervision also produces more informative target-model
rankings.

Monotonic recalibration reduces the English heads' mean Brier score to
\MeanETDCalibratedBrier{}. Because this transformation is strictly increasing
within each target-model head, it preserves the corresponding AUROC while
correcting systematic shifts in probability scale. Its improvement therefore
isolates the contribution of calibration. Nevertheless, the raw pooled heads
retain the lowest mean Brier score, and recalibration does not improve every
calibration diagnostic for every target model. Multilingual supervision
therefore contributes information that cannot be recovered by applying a
single post-hoc monotonic correction to English-trained predictions.

This distinction is important for routing. The policy compares predicted
success values across candidate models rather than ranking examples for one
model in isolation. Consequently, even an English-trained head with useful
within-model AUROC can produce poor decisions when its score scale is not
comparable with those of the other target-model heads.

\subsection{Reliability Diagrams}
\label{a}

The main text presents the reliability diagrams for \texttt{Qwen3-4B}. The
figures below report the corresponding per-language diagrams for the remaining
seven model configurations. Each pair compares the unchanged English-trained
probe with the equal-budget pooled multilingual probe at the model's frozen
shared reporting layer.

Across models, the English-trained diagrams show that calibration errors are
language dependent: predicted values that are appropriate for one language can
correspond to different observed success frequencies in another. This explains
why English transfer can preserve useful AUROC while obtaining worse ECE and
Brier scores. Pooled supervision generally moves the curves closer to the
diagonal and reduces their variation across languages, although it does not
eliminate all model- and language-specific calibration error.

The diagrams also show why a single macro-averaged calibration statistic is
insufficient on its own. Similar aggregate ECE values can arise from different
patterns, including systematic overprediction, systematic underprediction, or
opposing errors across languages that partially cancel when averaged. We
therefore interpret these figures together with AUROC, Brier score, NLL, and the
calibration coefficients reported in the complete tables.

\reliabilitycomparison{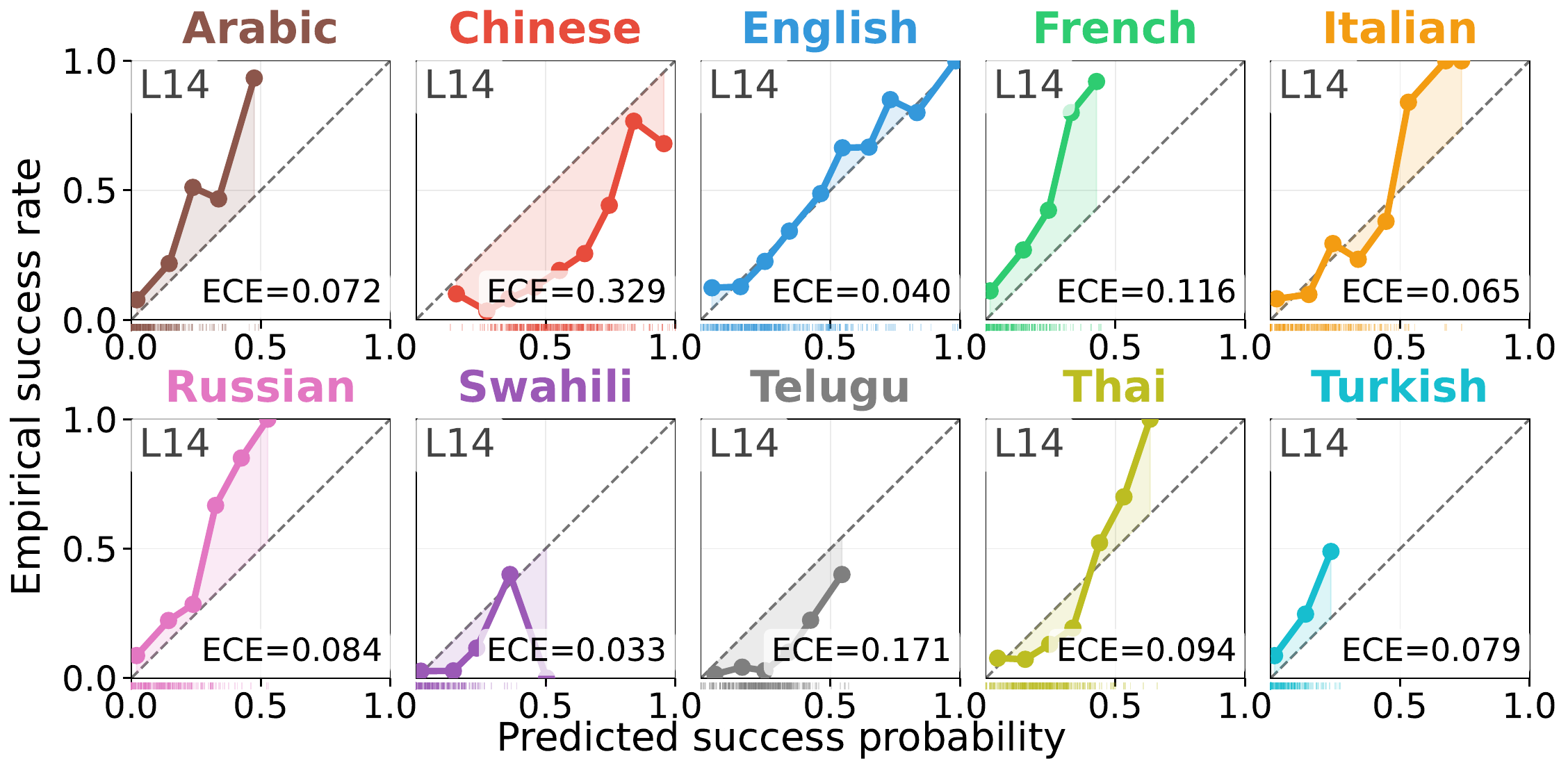}
\reliabilitycomparison{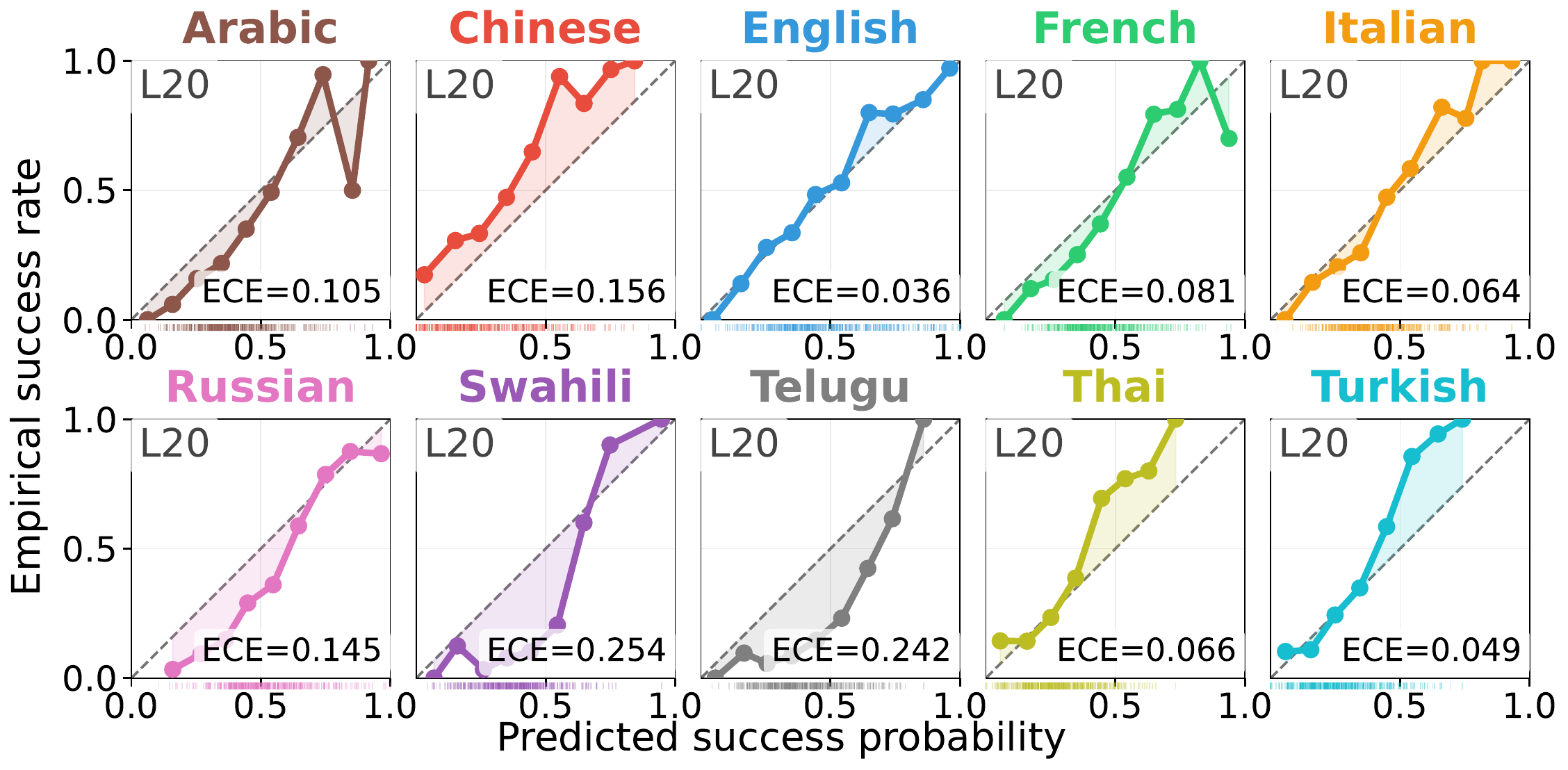}
\reliabilitycomparison{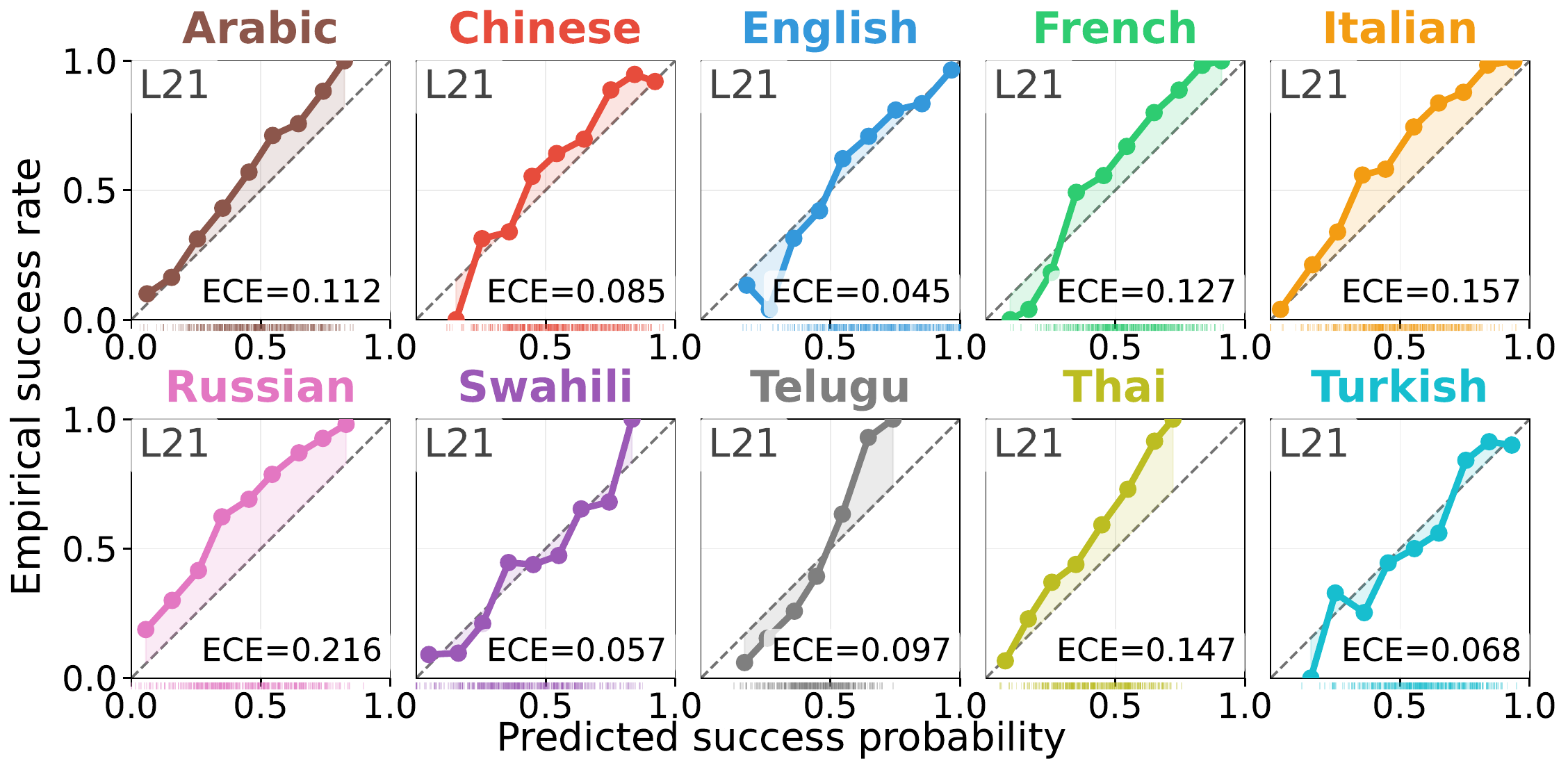}
\reliabilitycomparison{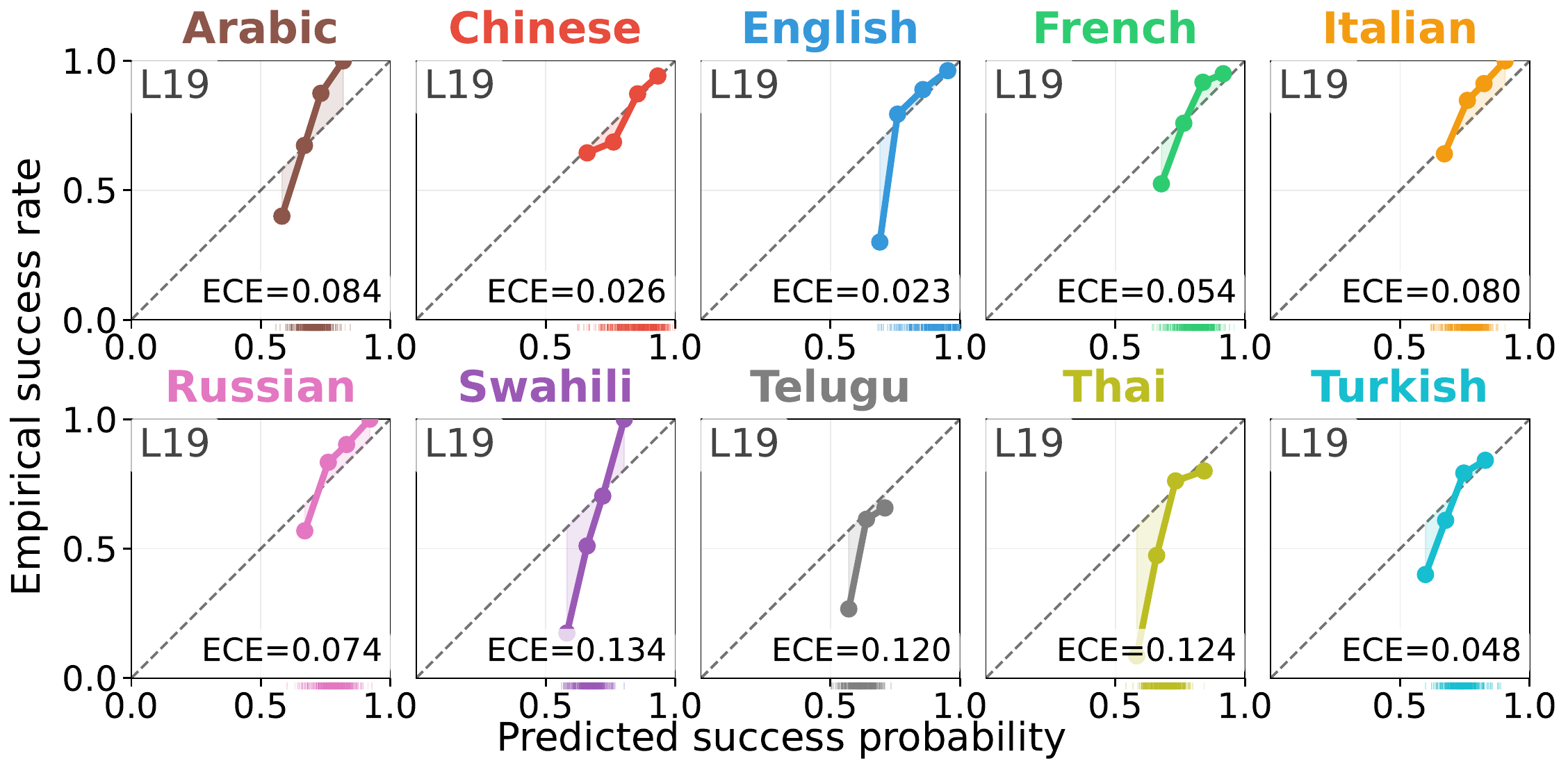}
\reliabilitycomparison{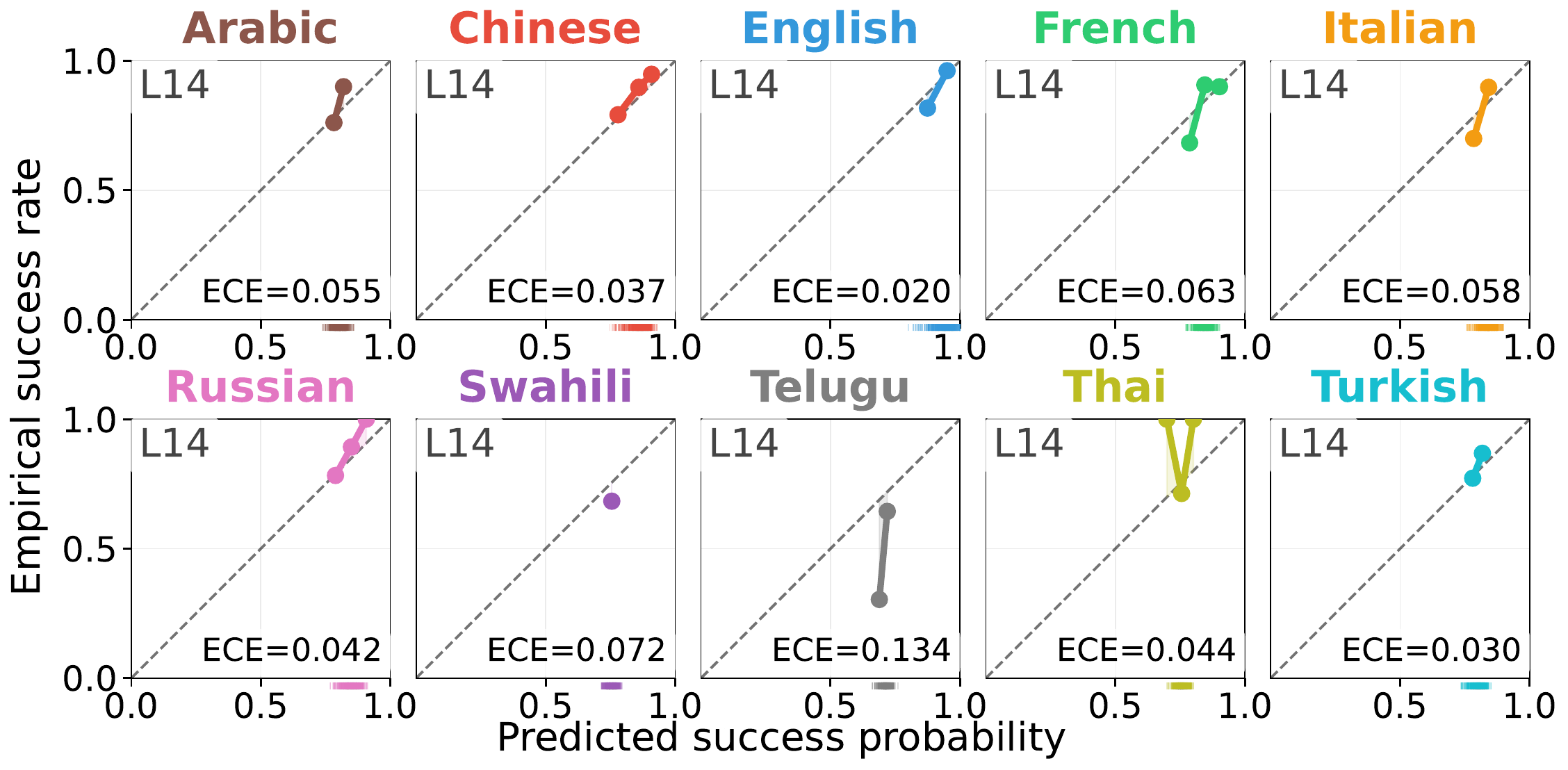}
\reliabilitycomparison{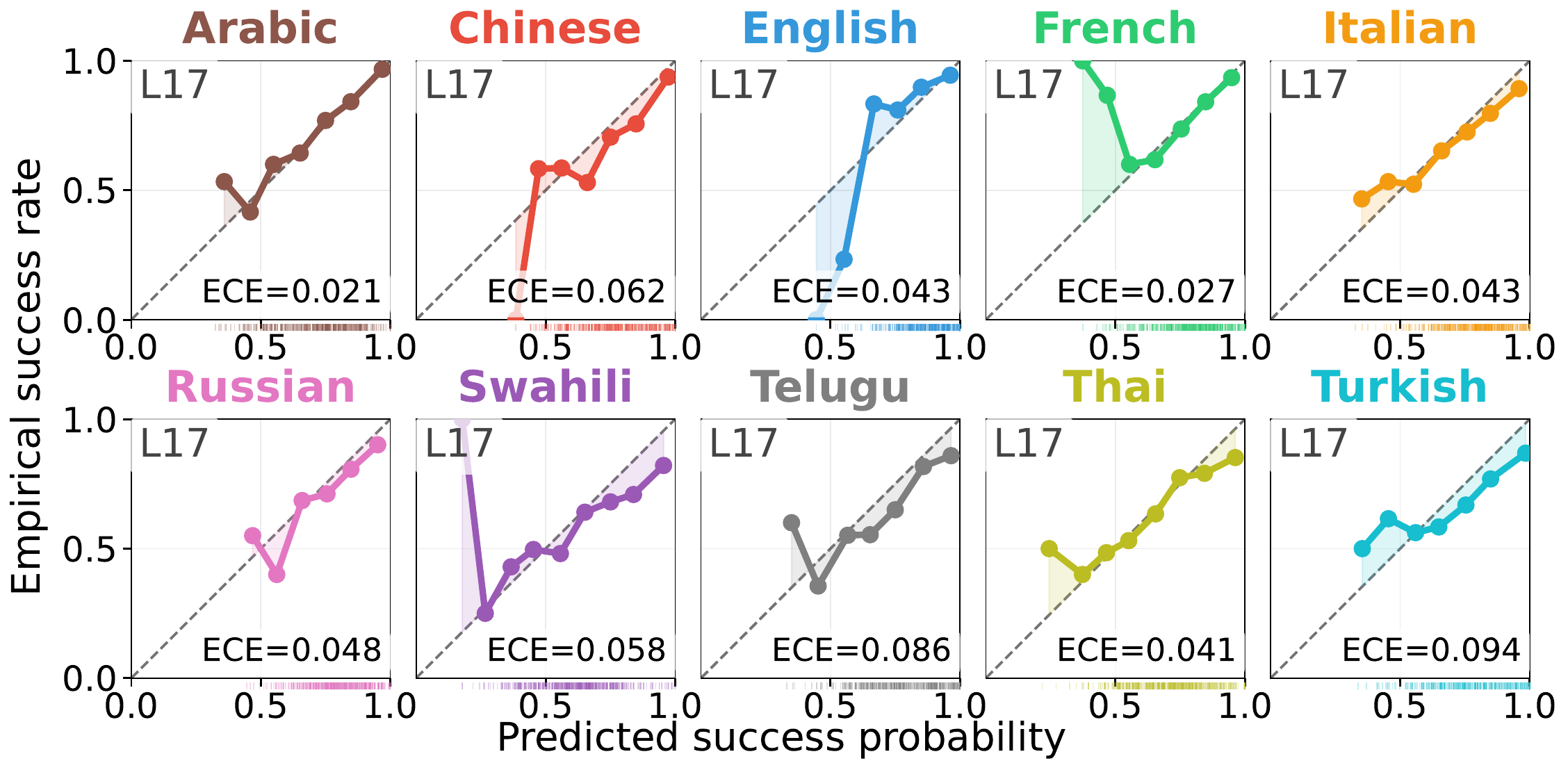}
\reliabilitycomparison{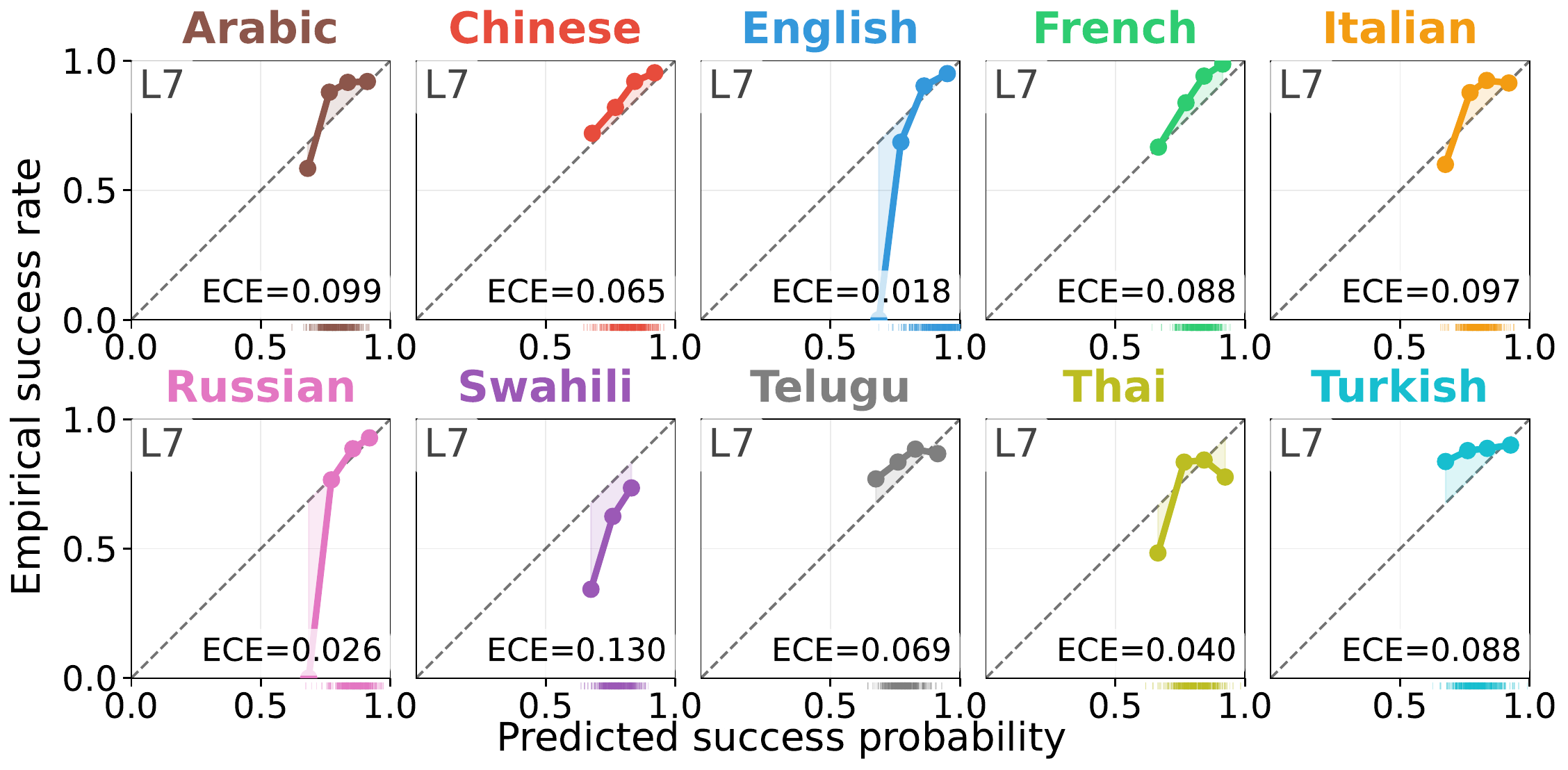}







\section{Monotonic Recalibration}
\label{a:monotonic-recalibration}

To isolate the effect of probability scaling, we recalibrate the predictions
of each English-trained ETD target-model head using the pooled multilingual
validation set. We fit a positive-slope Platt-scaling transformation
\citep{platt1999probabilistic}:
\begin{equation*}
    \widehat{p}^{\,\mathrm{cal}}
    =
    \sigma
    \left(
        a_m
        +
        \operatorname{softplus}(b_m)
        \operatorname{logit}
        \left(
            \widehat{p}^{\,\mathrm{raw}}
        \right)
    \right),
    \label{eq:monotonic-recalibration}
\end{equation*}
where \(a_m\) and \(b_m\) are fitted separately for each target model \(m\).
Predictions are clipped to
\([10^{-7},1-10^{-7}]\) before applying the logit.

The calibration parameters are estimated by minimizing binomial negative
log-likelihood on the nine non-English validation languages, using the observed
numbers of successful and unsuccessful rollouts. They are then frozen and
applied unchanged to the test predictions and routing simulation.

Because \(\operatorname{softplus}(b_m)>0\), the transformation is strictly
increasing. It therefore preserves the ordering of predictions within each
target-model head and cannot change its AUROC, apart from numerical ties.
However, since a different transformation is fitted for each target model,
recalibration can change the relative probability scales across models and
therefore affect routing decisions.

\end{document}